\documentclass[11pt]{article}
\usepackage[T1]{fontenc}
\usepackage[utf8]{inputenc}
\usepackage{graphicx,booktabs,tabularx,amsmath,amssymb,textcomp,xurl}
\usepackage[margin=1in]{geometry}
\usepackage[numbers,sort&compress]{natbib}
\usepackage[hidelinks]{hyperref}
\hypersetup{pdftitle={VGL-stGen: Learning a shared multimodal model for lung histology and spatial transcriptomics},pdfauthor={Haenara Shin, Dongjoo Lee, Sungwoo Bae, Kwon Joong Na, Daeseung Lee, Hongyoon Choi}}
\begin{document}
\title{VGL-stGen: Learning a shared multimodal model for lung histology and spatial transcriptomics}
\author{Haenara Shin\thanks{Email: \texttt{haenara.shin@portrai.io}.}, Dongjoo Lee, Sungwoo Bae,\\Kwon Joong Na, Daeseung Lee, Hongyoon Choi\thanks{Corresponding author. Email: \texttt{chy@portrai.io}.}\\[4pt]{\small Portrai, Inc., Seoul, Republic of Korea}}
\date{}
\maketitle
\begin{abstract}
Learning across histology and spatial transcriptomics requires a model that can connect tissue morphology, gene expression, and cellular identity within a shared framework. We developed VGL-stGen (Vision-Gene Language model–spatial transcriptomics generative), a multimodal large language model adapted from a pretrained medical vision-language model using large-scale paired lung H\&E and spatial transcriptomics data together with single-cell references. Training integrates image–expression and expression–cell-type pairs from multiple spatial transcriptomics and single-cell RNA-seq data, enabling complementary supervision across modalities. Gene-conditioned pseudo-labeling further transfers transcriptomic cell-type supervision to histology images. A single deployed VGL-stGen checkpoint supports thirteen evaluated task formulations spanning H\&E-based tissue and cell-type classification, gene-expression–based cell typing, gene prediction, and conditional expression completion from combinations of histology, cell identity, and observed gene expression. Task-specific decoding allows the same generative model to produce categorical, gene-list, and structured expression outputs. For expression completion, fixing queried gene identities and autoregressively predicting their expression levels substantially improves structured prediction, while controlled adaptation shows that restricting supervision to the requested expression-level tokens can further improve image-conditioned completion. Image-permutation experiments indicate that correctly paired histology provides additional information beyond observed gene-expression levels.

External classification relies partly on computational references with shared annotation lineage, and some H\&E sources were exposed during cancer training or model development. Within these constraints, VGL-stGen demonstrates that a medical vision-language backbone can be adapted with large-scale spatial transcriptomics and histology data into a unified multimodal model capable of performing diverse H\&E- and gene-expression–related tasks. This framework provides a foundation for connecting tissue morphology and spatial molecular states through shared generative modeling.
\end{abstract}
\section{Introduction}
Recent advances in spatial biology have enabled computational models to learn representations from histology images and spatially resolved gene expression. However, most existing models are designed around specific modalities or prediction tasks, leaving histological morphology, gene expression, and cellular identity largely separated across modeling frameworks. A unified model that can learn the underlying biology shared across these modalities and flexibly reason over different combinations of images, genes, and cell states remains underexplored. To address this gap, we developed VGL-stGen, a multimodal large language model trained on large-scale lung and lung cancer spatial transcriptomics data. We assembled a lung training corpus combining paired histology and spatial transcriptomics from a total of 69 samples (47 Xenium samples and 22 Visium samples) in PortraiATLAS with single-cell references and cancer annotations. By integrating these relationships within a shared vision-gene-language framework, VGL-stGen is designed to learn across histological and molecular representations and perform diverse tasks from different combinations of H\&E images, gene expression, and cellular information.

Existing image–omics methods use different observations and target representations. ST-Net predicts expression from histology \citep{r28}, BLEEP retrieves reference expression through contrastive embeddings \citep{r39}, and GenAR generates expression counts autoregressively \citep{r41}. SpaCKLE completes expression using neighbouring spots \citep{r40}. OmiCLIP aligns histology with ranked gene text for multiple tasks \citep{r29}, and SPADE learns aligned histology-expression representations \citep{r42}. STPath integrates histology, observed expression and spatial context for generative prediction and spot imputation \citep{r43}. GHIST predicts single-cell expression from histology using auxiliary cell-type and neighbourhood-composition tasks \citep{r44}. These methods establish the importance of specifying which molecular observations are available at inference and which targets are predicted.

VGL-stGen adapts the pretrained medical vision-language model MedGemma-1.5-4b-it \citep{r01} for shared conditional generation across the supported lung tasks. Gene-conditioned pseudo-labels supply image cell-type supervision. The model generates gene lists and uses constrained decoding for categorical labels and queried expression levels. For conditional completion, the input includes 75 observed levels and either histology or a supplied cell type; the output contains levels for 25 named genes. Fixing the query names separates supplied identifiers from predicted values. The distinction motivates a controlled comparison of full-answer supervision with loss masking over the requested level tokens.

We assess image cell typing before and after pseudo-label training, evaluate fixed-name decoding from the deployed weights, and compare expression supervision under matched initial weights, training examples and update budgets. Image permutation tests whether correct pairing contributes when observed gene levels remain available. Task-specific classifiers and full-input MLPs provide reference performance with the corresponding observed inputs. The deployed checkpoint’s experiments establish task-dependent completion effects.

\section{Methods}
\subsection{Data and shared-model learning}
Task instructions condition a shared autoregressive decoder on the supported input combinations (Fig. 1). Outputs are categorical labels, gene lists or discrete expression levels. The two expression-completion tasks predict 25 named gene levels from 75 observed levels together with either histology or a supplied cell type.

\begin{figure}[!htbp]
\centering
\includegraphics[width=122.000mm]{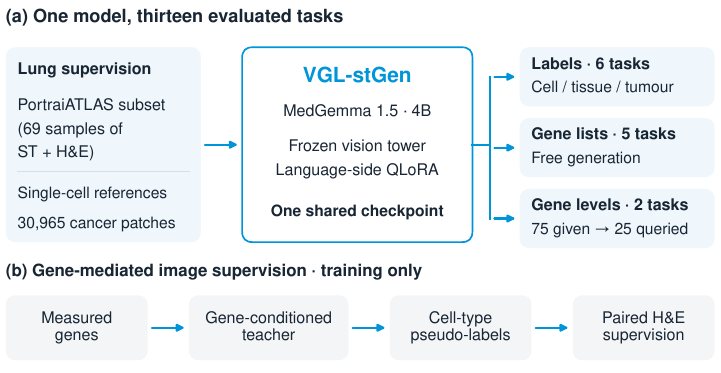}
\caption{VGL-stGen overview. (a) Complementary lung supervision adapts one medical vision–language model for 13 evaluated tasks. The 6 label, 5 gene-list and 2 expression-level tasks share a decoder and checkpoint. (b) Gene-conditioned cell-type pseudo-labels supervise paired images during training. Conditional expression completion uses 75 observed levels and 25 named queries. Detailed task inputs and training stages are specified in the Methods and supplementary material.}
\label{fig:overview}
\end{figure}
Training integrates paired histology and spatial transcriptomics from 69 ST–H\&E samples within a subset of the PortraiATLAS database, together with single-cell references and 30,965 unique cancer patches. Spatial-transcriptomic and pathologist-derived H\&E patches were extracted at a nominal 112-\ensuremath{\mu}m field of view and stored at 224 \ensuremath{\times} 224 pixels. LubLung training crops contained approximately 87 \ensuremath{\mu}m of tissue and were resized and padded to the same pixel dimensions. Masked-expression training data cover 2,000 genes for the image task and 1,478 for the cell-type task. Targets are ‘Absent’ for undetected expression and four quartile levels: ‘Low’, ‘Med’, ‘High’ and ‘Very High’, defined from detected expression within a cell or Visium spot.

The \texttt{medgemma-1.5-4b-it} backbone has a frozen vision tower and QLoRA-adapted language projections (rank 16, alpha 32, 4-bit NormalFloat (NF4) weights, bfloat16 computation) \citep{r07,r08}. Training applies autoregressive supervision to the target answer tokens of task-specific image–text or text-only examples. Transcriptomic adaptation includes gene-list, cell-type and expression-bin objectives. The teacher predicts cell type from measured genes; its highest-likelihood label supervises the paired image. A top-two log-likelihood gap of at least 1.0 retains 1.32M of 1.75M pseudo-labelled items. The student trains on 300K image-task and 200K replay rows for one epoch at learning rate 1e-4.

Final training adds binary tumour and nine-class tissue objectives with nine transcriptomic replay objectives, excluding image-to-cell-type supervision. The joint image-and-genes cell-type task has no direct training objective. We average six late rank-16 adapters \citep{r30}, representing their mean at rank 96 through concatenation. The averaging window was selected using the 39-slide TCGA development subset. For the main benchmarks, binary tumour scoring uses merged bfloat16 weights; nine-class tissue, cell-type, gene-list and expression scoring use the NF4-base adapter.

Table 1 specifies the evaluated inputs, outputs and conditions.

\begin{table}[!htbp]
\centering\small
\caption{13 evaluated tasks supported by the deployed checkpoint. BA: balanced accuracy; F1m: macro-F1; F1g: gene-set F1; F1p: pair-F1; Acc: accuracy; AR: autoregressive. H\&E overlap denotes partial exposure of evaluation tissue during cancer training. Gene-list completion supplies 15 names and predicts five; expression completion supplies 75 levels and queries 25 named genes. Task 10 supplies true answer length; task 13 supplies ten gene–level pairs. The complete score inventory is supplementary.}
\label{tab:task_definitions}
\begin{tabularx}{\textwidth}{@{}>{\hsize=0.240\hsize\raggedright\arraybackslash}X>{\hsize=1.960\hsize\raggedright\arraybackslash}X>{\hsize=0.520\hsize\raggedright\arraybackslash}X>{\hsize=1.280\hsize\raggedright\arraybackslash}X@{}}
\toprule
ID & Input \ensuremath{\rightarrow} output & Metric & Evaluation condition \\
\midrule
1 & Genes \ensuremath{\rightarrow} cell type & BA / F1m & External \\
2 & Image + genes \ensuremath{\rightarrow} cell type & BA & H\&E overlap \\
3 & Image \ensuremath{\rightarrow} cell type & BA & H\&E overlap \\
4 & Image \ensuremath{\rightarrow} tumour & BA & 452 slides \\
5 & Image \ensuremath{\rightarrow} tissue & F1m & 39-slide pool \\
6 & Image \ensuremath{\rightarrow} genes & F1g & H\&E overlap \\
7 & Cell type \ensuremath{\rightarrow} genes & F1g & External \\
8 & Image + partial \ensuremath{\rightarrow} genes & F1g & H\&E overlap \\
9 & Cell type + partial \ensuremath{\rightarrow} genes & F1g & External \\
10 & Image \ensuremath{\rightarrow} very-high genes & F1g & Internal; true length \\
11 & Image + levels \ensuremath{\rightarrow} levels & F1p & Internal AR \\
12 & Cell type + levels \ensuremath{\rightarrow} levels & F1p & Internal AR \\
13 & Levels \ensuremath{\rightarrow} cell type & Acc & Internal \\
\bottomrule
\end{tabularx}
\end{table}
\subsection{Structured prediction}
Closed labels use prefix-tree decoding; twenty-gene inputs preserve expression rank. For completion, a parser extracts query names from the 100-gene input. With names fixed, the model predicts each level from five permitted first tokens, conditioned autoregressively on its own preceding answers. Pair-F1 equals accuracy over the queried gene–level pairs.

\subsection{Evaluation cohorts and reference annotations}
Table 2 summarises evaluation units and annotation provenance.

\begin{table}[!htbp]
\centering\small
\caption{Evaluation units and reference provenance, with spatial datasets grouped by platform. Visium units are spots; Xenium units are cells. Counts refer to processed evaluation items rather than training samples. Source-level provenance and the different exposure of the two Visium sections are retained in the supplementary methods. TCGA-LUAD denotes lung adenocarcinoma. The tissue-classification analysis uses the 39-slide development subset.}
\label{tab:cohorts}
\begin{tabularx}{\textwidth}{@{}>{\hsize=0.840\hsize\raggedright\arraybackslash}X>{\hsize=0.690\hsize\raggedright\arraybackslash}X>{\hsize=1.470\hsize\raggedright\arraybackslash}X@{}}
\toprule
Cohort & Evaluated units & Reference / exposure \\
\midrule
Visium & 9,964 spots / 2 sections & Transferred cell types; one section supplied H\&E for cancer training \\
Xenium & 214,246 cells / 2 sections & Transferred cell types; both sections supplied H\&E for cancer training \\
TCGA-LUAD binary & 357,692 patches / 452 slides / 398 patients & ARA-CNN labels; includes development subset \\
TCGA tissue classification & 30,719 patches / 39 slides & Development subset; pooled transductive correction \\
Masked expression benchmark & 1,000 rows/task & Internal; 7 image samples / 1 cell-type sample identifier \\
\bottomrule
\end{tabularx}
\end{table}
The external spatial and TCGA classification benchmarks use computationally derived reference annotations (often termed silver labels). ARA-CNN cancer segmentations \citep{r26} share LubLung lineage with training; TACCO cell-type transfers \citep{r35} use the training lung cancer atlas (LuCA) \citep{r32}. Scores measure agreement with these annotations, which can preserve shared errors. The four public 10x sections \citep{r31}, excluded from transcriptomic training, pair H\&E with measured expression: 214,246 Xenium cells and 9,964 Visium spots support centre-cell and dominant regional typing, respectively. Gene-list and expression targets derive from measured expression. External cell-type-conditioned gene queries use transferred labels; internal cell-type-conditioned expression queries use LuCA atlas labels. TCGA-LUAD supplies cancer images \citep{r36}.

Three of the four spatial sections supplied 4,846 binary cancer training rows. The Visium squamous cell carcinoma (SCC) section has no source match in that cancer-training manifest. Image-conditioned external tasks therefore include previously seen tissue with unseen task labels. A 39-slide TCGA subset informed development and is included in the 452-slide evaluation. Sample and recorded cell/spot identifiers are disjoint within the expression-task training and test manifests. However, all seven samples in the masked-image benchmark were represented in gene-list training, including 692 of the 1,000 benchmark cells or spots. This internal benchmark therefore does not establish patient-independent generalisation across the full training sequence.

\subsection{Matched comparisons and statistical analysis}
We compare the deployed VGL-stGen checkpoint with ten off-the-shelf models: MedGemma-1.5-4B \citep{r01}, Gemma3-4B \citep{r47}, Gemma4-12B and Gemma4-31B \citep{r48}, Qwen3-VL-8B and Qwen3-VL-32B \citep{r49}, Qwen3.8-27B \citep{r50}, DeepSeek-VL2-small and DeepSeek-VL2 \citep{r51}, and GLM-4.6V-Flash \citep{r52}. All receive the same task inputs and instructions, with the original task-specific generation limits, batch composition, output parser and metric. Native chat templates, tokenizers and image processors are retained. Task 4 uses BF16; all other tasks use NF4 double quantization with BF16 computation in quantized layers. The deployed VGL-stGen adapter and merged weights follow this same task-specific assignment. This comparison evaluates lung-specific adaptation against pretrained multimodal models; task-specific trained classifiers and MLPs are assessed separately.

Categorical tasks use allowed-label decoding, gene-list tasks use generation followed by the original gene parser, and expression completion predicts the levels of fixed queried genes autoregressively. These are structured prediction protocols rather than a uniform test of unrestricted JSON generation. A supplementary format-constrained arm repeats tasks 6–10 and 13 for all eleven models under the same budgets. It restricts gene-array syntax or permitted label spellings without supplying the correct genes or restricting predictions to an assay panel. Paired bootstrap intervals use 1,000 draws: cells or spots for tasks 1–3, slides for task 4, patches for task 5, and complete input rows for tasks 6–13. These unadjusted intervals condition on fitted models and do not establish generalisation across independent patients or sections. Detailed settings, model revisions and all paired intervals accompany the supplementary analysis.

Gene comparators receive the same ranked twenty-gene lists; comparators include logistic regression, Geneformer V2 and C2S variants, with language-model controls \citep{r09,r45,r10,r11,r12}. Morphology probes use fifteen frozen encoders with class-balanced logistic regression (C=1). Transferred probes learn gene-teacher labels; target-cohort references use five-fold stratified cell-level cross-validation on the 54,058-item subset. Fourteen cancer encoders \citep{r13,r14,r15,r16,r17,r18,r19,r20,r21,r22,r23,r24,r25} carry binary probes trained on the same 30,965 unique cancer patches as VGL-stGen, which also receives auxiliary supervision.

The conditional-expression image benchmark spans seven sample identifiers: three Xenium (665 cells) and four Visium (335 spots). Task-specific cell-type comparator intervals use section-then-item paired bootstrap (2,000 draws). Primary cancer intervals resample slides; development-patient exclusion resamples patients (1,000 draws each). New expression contrasts use 2,000 paired row or sample draws within platform, preserving platform proportions. The cell-type-conditioned benchmark has one sample identifier, permitting only row intervals. All 95\% confidence intervals (CIs) condition on fitted models, without multiplicity adjustment; sample identifiers are not verified patients. For VGL-stGen, standard deviation (SD) is computed across three additional-training runs from the same initial checkpoint.

\subsection{Application analyses}
We evaluated genes \ensuremath{\rightarrow} cell type (Task 1) with the checkpoint (merged bfloat16 weights, constrained decoding) on two application sources. The first was GSE127465, a single-cell RNA-seq dataset of seven treatment-naive NSCLC patients with published annotations \citep{r46}. We used 2,000 highly variable genes. Each of 7,993 randomly sampled tumor-tissue cells was then represented by its 20 highest-expressed genes in rank order. To match the authors' 11 major types, we merged macrophage/monocyte and cDC predictions into their MoMacDC class. Patient-specific malignant clusters were treated as epithelial, and we also report results without them.

The second source was the public 10x Xenium Human Lung Cancer FFPE section (377-gene panel) \citep{r31}. We scored 12,000 random cells with at least 10 transcripts, using the same top-20 input. Because this section has no published annotation, we crosstabulated each predicted label against measured expression of canonical marker sets. These application conditions differ from the main 224,210-item cell-type benchmark; the supplementary methods specify reference construction.

For the image \ensuremath{\rightarrow} genes task, we analysed 459 TCGA-LUAD patients with a readable diagnostic FFPE slide and open primary-tumour RNA-seq \citep{r36}. This cohort has no patients in common with the TCGA development subset. The model returned 20 gene symbols for each of 112 random 128 \ensuremath{\mu}m tissue patches per slide. A slide's value for a gene is the fraction of its patches naming that gene. We compared this value with log2(TPM+1) for 317 genes, restricted to genes named in at least 10\% of slides and expressed (TPM > 1) in at least half of cases. Both matrices were gene-centred to focus comparisons on patient-to-patient variation.

The endpoints were per-gene Pearson correlation across patients (Benjamini–Hochberg FDR, compared with a patient-permutation null) and correlation of marker-programme scores. Patient matching was summarized by the median percentile at which each slide ranks its own bulk profile among all 459 patients. Controls were a shuffled slide–bulk pairing, restriction to each tissue source site, and incremental cross-validated R² after sequencing depth, magnification, source site and tumor fraction. TCGA slides and RNA come from different tissue blocks.

\subsection{Analyses of learning and input contributions}
We assess cell typing on images before and after gene-mediated supervision. Separate expression analyses compare full-answer and level-token supervision from the same deployed checkpoint, using 4,096 additional rows, 256 updates and three matched seeds. Image permutation retains observed expression and queried genes. Full-input multilayer perceptrons (MLPs) receive the same molecular context, with image features or the supplied cell type; their encoders and training histories differ. These analyses test specific learning and input choices. Their checkpoints and evaluation protocols are detailed in the supplementary material.

\section{Results}
\subsection{Shared-model task coverage and multimodal comparisons}
The deployed checkpoint supports the thirteen tasks defined in Table 1. Table 3 and Table 4 compare it with ten off-the-shelf medical and general-purpose multimodal models under matched task-specific protocols.

\begin{table}[!htbp]
\centering\scriptsize\setlength{\tabcolsep}{2pt}
\caption{Medical and general-purpose multimodal comparisons on tasks 1–5. BA is balanced accuracy and F1m is macro-F1. T1–3 use 224,210 items each; T4 uses 357,692 patches from 452 slides; T5 uses 30,719 patches from 39 development slides. T4 uses BF16 and reports slide-bootstrap means; the other tasks use NF4. T5 shows raw \ensuremath{\rightarrow} pooled-corrected F1m for every model. Native image processors are retained. Task key: T1, genes \ensuremath{\rightarrow} cell type; T2, image + genes \ensuremath{\rightarrow} cell type; T3, image \ensuremath{\rightarrow} cell type; T4, image \ensuremath{\rightarrow} binary tumor status; T5, image \ensuremath{\rightarrow} 9-class tissue type.}
\label{tab:vlm_classification}
\begin{tabularx}{\textwidth}{@{}>{\hsize=1.500\hsize\raggedright\arraybackslash}X>{\hsize=1.080\hsize\raggedright\arraybackslash}X>{\hsize=0.600\hsize\raggedright\arraybackslash}X>{\hsize=0.600\hsize\raggedright\arraybackslash}X>{\hsize=0.660\hsize\raggedright\arraybackslash}X>{\hsize=1.560\hsize\raggedright\arraybackslash}X@{}}
\toprule
Model & T1 BA / F1m & T2 BA & T3 BA & T4 BA & T5 F1m \\
\midrule
VGL-stGen & 0.6293 / 0.6023 & 0.5394 & 0.2012 & 0.7524 & 0.1200 \ensuremath{\rightarrow} 0.2801 \\
MedGemma-1.5-4B & 0.1836 / 0.1576 & 0.2070 & 0.1105 & 0.5954 & 0.0911 \ensuremath{\rightarrow} 0.1711 \\
Gemma3-4B & 0.2254 / 0.1738 & 0.1853 & 0.1250 & 0.6637 & 0.1334 \ensuremath{\rightarrow} 0.1347 \\
Qwen3-VL-8B & 0.4662 / 0.3148 & 0.4572 & 0.0995 & 0.5182 & 0.1102 \ensuremath{\rightarrow} 0.1630 \\
DeepSeek-VL2-small & 0.1314 / 0.1140 & 0.1380 & 0.1100 & 0.5351 & 0.0649 \ensuremath{\rightarrow} 0.1321 \\
GLM-4.6V-Flash & 0.3444 / 0.2775 & 0.3160 & 0.1008 & 0.5808 & 0.1411 \ensuremath{\rightarrow} 0.1733 \\
Gemma4-12B & 0.4023 / 0.3684 & 0.4071 & 0.1440 & 0.6986 & 0.1651 \ensuremath{\rightarrow} 0.1763 \\
Qwen3.8-27B & 0.3414 / 0.2703 & 0.3302 & 0.1265 & 0.6974 & 0.1947 \ensuremath{\rightarrow} 0.2613 \\
Gemma4-31B & 0.5035 / 0.4870 & 0.4634 & 0.1652 & 0.6719 & 0.1887 \ensuremath{\rightarrow} 0.1958 \\
DeepSeek-VL2 & 0.1510 / 0.1035 & 0.1595 & 0.1103 & 0.5983 & 0.0554 \ensuremath{\rightarrow} 0.1563 \\
Qwen3-VL-32B & 0.4822 / 0.3688 & 0.4739 & 0.1110 & 0.6742 & 0.1625 \ensuremath{\rightarrow} 0.1953 \\
\bottomrule
\end{tabularx}
\end{table}
\begin{table}[!htbp]
\centering\scriptsize\setlength{\tabcolsep}{2pt}
\caption{Multimodal comparisons on gene and expression tasks, with 1,000 rows per task and NF4 weights. T6–10: gene-set F1; T11–12: pair-F1; T13: accuracy. T10 supplies the true answer length; T11–12 supply 75 observed levels and query 25 named genes. T6–10 and T13 use the primary generation protocol, separately from the format-constrained control. Values preserve the original score aggregation and rounding. Task key: T6, image \ensuremath{\rightarrow} gene names; T7, cell type \ensuremath{\rightarrow} gene names; T8, image + partial gene list \ensuremath{\rightarrow} missing gene names; T9, cell type + partial gene list \ensuremath{\rightarrow} missing gene names; T10, image \ensuremath{\rightarrow} very-high-expression gene names; T11, image + observed levels \ensuremath{\rightarrow} queried levels; T12, cell type + observed levels \ensuremath{\rightarrow} queried levels; T13, gene-level pairs \ensuremath{\rightarrow} cell type.}
\label{tab:vlm_molecular}
\begin{tabularx}{\textwidth}{@{}>{\hsize=2.160\hsize\raggedright\arraybackslash}X>{\hsize=0.855\hsize\raggedright\arraybackslash}X>{\hsize=0.855\hsize\raggedright\arraybackslash}X>{\hsize=0.855\hsize\raggedright\arraybackslash}X>{\hsize=0.855\hsize\raggedright\arraybackslash}X>{\hsize=0.855\hsize\raggedright\arraybackslash}X>{\hsize=0.855\hsize\raggedright\arraybackslash}X>{\hsize=0.855\hsize\raggedright\arraybackslash}X>{\hsize=0.855\hsize\raggedright\arraybackslash}X@{}}
\toprule
Model & T6 & T7 & T8 & T9 & T10 & T11 & T12 & T13 \\
\midrule
VGL-stGen & 0.1173 & 0.0965 & 0.0604 & 0.0484 & 0.2828 & 0.6596 & 0.6858 & 0.7420 \\
MedGemma-1.5-4B & 0.0000 & 0.0005 & 0.0000 & 0.0000 & 0.0003 & 0.5733 & 0.5540 & 0.0000 \\
Gemma3-4B & 0.0088 & 0.0345 & 0.0108 & 0.0000 & 0.0058 & 0.5072 & 0.4996 & 0.3500 \\
Qwen3-VL-8B & 0.0000 & 0.0864 & 0.0210 & 0.0380 & 0.0032 & 0.1634 & 0.2372 & 0.3500 \\
DeepSeek-VL2-small & 0.0000 & 0.0080 & 0.0026 & 0.0000 & 0.0042 & 0.1600 & 0.2165 & 0.0630 \\
GLM-4.6V-Flash & 0.0044 & 0.0397 & 0.0140 & 0.0237 & 0.0030 & 0.3662 & 0.3627 & 0.4260 \\
Gemma4-12B & 0.0248 & 0.0791 & 0.0140 & 0.0463 & 0.0071 & 0.2692 & 0.2077 & 0.3960 \\
Qwen3.8-27B & 0.0269 & 0.0699 & 0.0209 & 0.0178 & 0.0131 & 0.6130 & 0.5621 & 0.4560 \\
Gemma4-31B & 0.0556 & 0.1003 & 0.0382 & 0.0502 & 0.0229 & 0.4931 & 0.5108 & 0.4620 \\
DeepSeek-VL2 & 0.0005 & 0.0282 & 0.0000 & 0.0105 & 0.0000 & 0.1043 & 0.1127 & 0.2950 \\
Qwen3-VL-32B & 0.0197 & 0.0839 & 0.0022 & 0.0452 & 0.0047 & 0.4315 & 0.3790 & 0.4020 \\
\bottomrule
\end{tabularx}
\end{table}
For gene-based cell typing, VGL-stGen reaches balanced accuracy 0.6293 and macro-F1 0.6023, compared with 0.5035 and 0.4870 for Gemma4-31B. Adding an image to the gene input yields balanced accuracy 0.5394, versus a comparator maximum of 0.4739 for Qwen3-VL-32B. Image-only cell typing remains harder: VGL-stGen scores 0.2012 versus 0.1652 for Gemma4-31B. The positive differences quantify performance against the computational reference annotations under the stated exposure conditions.

Binary tumour balanced accuracy is 0.7524 versus 0.6986 for Gemma4-12B. The nine-class tissue result depends on readout: VGL-stGen's raw macro-F1 is 0.1200, below Qwen3.8-27B's 0.1947, whereas pooled probability correction yields 0.2801 versus 0.2613. The correction uses the evaluation pool and was selected during development; both scores are therefore reported.

Expression-bin prediction is a core component of the deployed model. Image-conditioned expression completion reaches pair-F1 0.6596, versus 0.6130 for Qwen3.8-27B; cell-type-conditioned completion reaches 0.6858 versus 0.5621. These results use 75 observed levels and 25 queried genes. On the two cell-type-conditioned gene-list tasks, Gemma4-31B has slightly higher point estimates (0.1003 versus 0.0965 and 0.0502 versus 0.0484), with paired intervals spanning zero. Very-high-gene prediction scores 0.2828 under the supplied true-length condition, while levels-to-cell-type accuracy is 0.7420. The latter exceeds the evaluation-set majority-label baseline of 0.3620.

The format-constrained evaluation retains VGL-stGen's reported scores and improves several comparator outputs. The strongest comparator scores are 0.0611 versus VGL-stGen's 0.1173 for image-to-genes prediction, 0.0244 versus 0.2828 for very-high-gene prediction, and 0.4720 versus 0.7420 for levels-to-cell-type classification. These controls support predictive gains beyond output syntax alone, while retaining the fixed-budget qualification for models that generate intermediate reasoning.

\subsection{Task-specific classification benchmarks}
For genes-to-cell-type, VGL-stGen achieves balanced accuracy 0.6293 and macro-F1 0.6023. Rank-weighted logistic regression gives 0.6616 and 0.5560, respectively. VGL-stGen's macro-F1 margin is +0.0463 with section-paired CI [+0.0276, +0.0597], while logistic regression has higher balanced accuracy. Fig. 2 shows the two fine-tuned Geneformer variants separately.

\begin{figure}[!htbp]
\centering
\includegraphics[width=137.000mm]{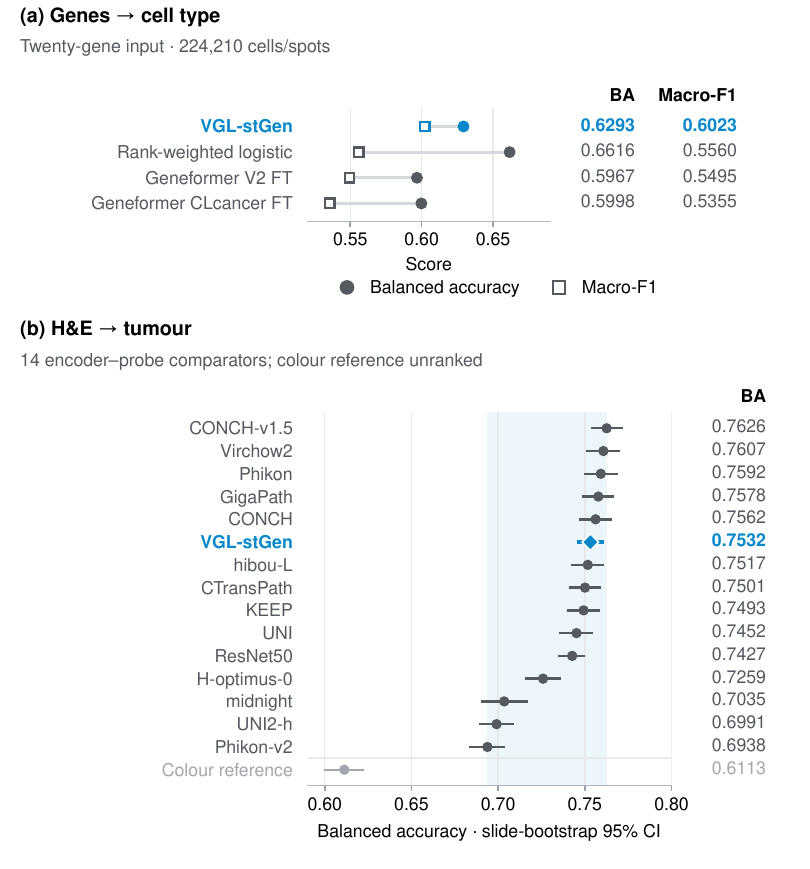}
\caption{Task-specific comparisons. (a) Balanced accuracy (BA; circles) and macro-F1 (squares) for the same 20-gene input; connecting lines pair metrics, not confidence intervals. FT denotes fine-tuning; the two Geneformer variants are separate systems. (b) 14 encoder–probe comparators, VGL-stGen and an unranked colour reference on 452 TCGA slides. Intervals resample slides; the shaded band spans the 14 encoder point estimates. Classification references are computationally derived. TCGA includes the 39-slide development subset; point ordering does not establish significance between neighbouring models.}
\label{fig:comparisons}
\end{figure}
On 357,692 patches from 452 TCGA slides, VGL-stGen achieves balanced accuracy 0.7532 [0.7452, 0.7610] (bootstrap mean) and AUROC 0.8266 against ARA-CNN-derived references. Fourteen encoder–probe comparators span 0.6938–0.7626; VGL-stGen ranks sixth of fifteen systems. Its frozen-tower probe scores 0.7522. Excluding all 40 slides belonging to the 39 development patients leaves 412 slides, 359 patients and 326,118 patches: observed balanced accuracy is 0.7523 versus 0.7534 on the full set, patient CI [0.7407, 0.7628], with rank unchanged.

\subsection{Contributions of cross-modal supervision and expression inputs}
Gene-mediated supervision increases image-to-cell-type balanced accuracy from 0.161 to 0.226; the deployed checkpoint scores 0.2012 after joint cancer training and averaging. This supports transferring cell-type supervision through measured genes while retaining the final checkpoint's performance as the primary endpoint.

In the separate three-seed expression study, level-token supervision increases image-conditioned pair-F1 from 0.6720 to 0.6805. Cell-type-conditioned pair-F1 is 0.7201, compared with 0.7221 for full-answer supervision. Correct image pairing contributes when observed expression is retained: permutation reduces mean pair-F1 by 0.00776 under level-token supervision. Full-input MLPs retain higher point estimates on both completion tasks. Supplementary methods, tables and figures report both supervision conditions, paired intervals, secondary metrics and input controls. These additional-training checkpoints are separate from the deployed model evaluated across thirteen tasks.

\subsection{Gene-based cell typing and spatial application}
On GSE127465, predicted cell types recovered the published annotation with balanced accuracy 0.815 and macro-F1 0.819 across 11 classes. Balanced accuracy was 0.814 when malignant cells were excluded, and 0.70–0.86 in individual patients (Fig. 3(a–c)). The input format mattered. Shuffling the order of the same 20 genes lowered balanced accuracy to 0.775, and omitting the gene filter lowered it to 0.282.

\begin{figure}[!htbp]
\centering
\includegraphics[width=140.000mm]{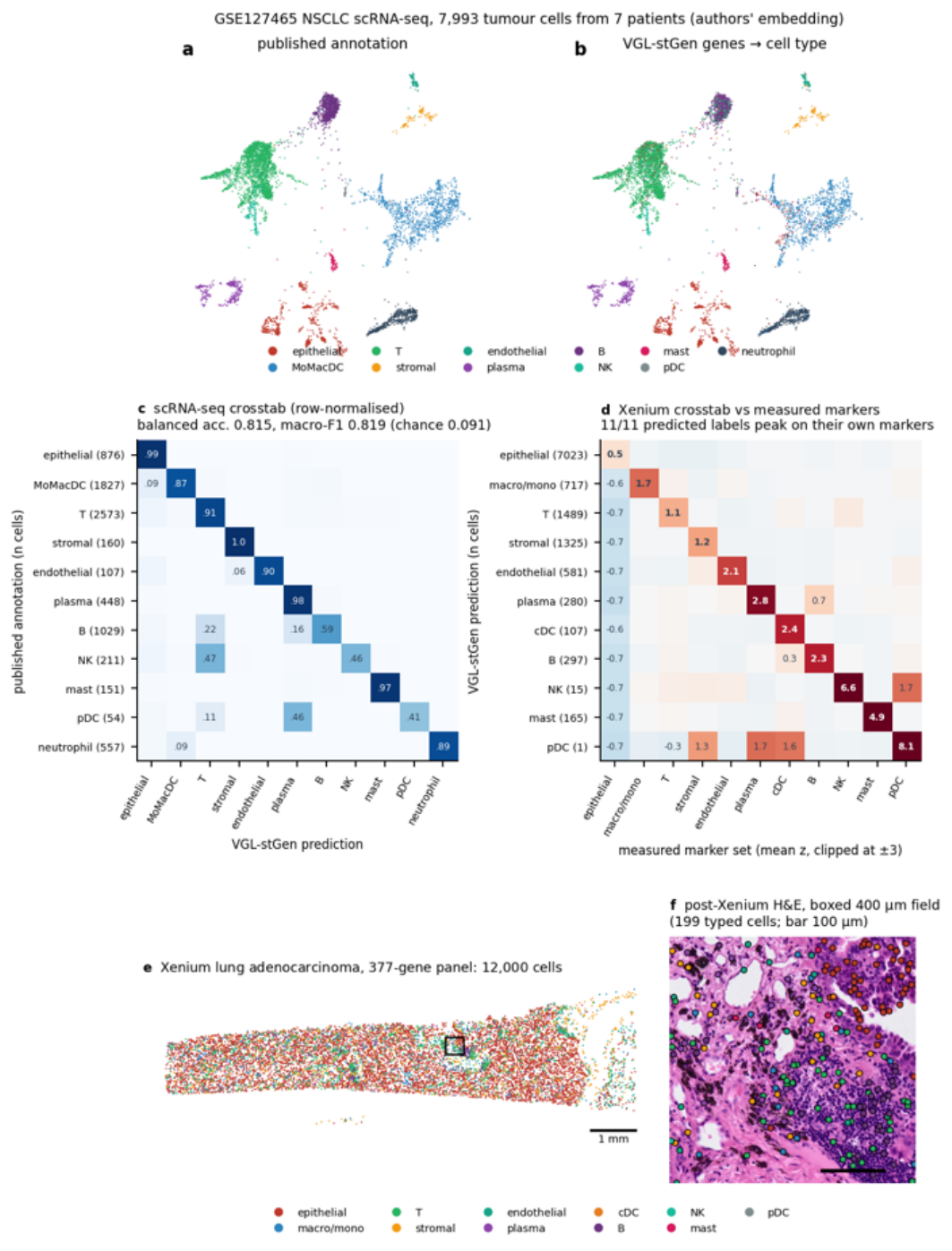}
\caption{Gene-conditioned cell typing. (a,b) Published and predicted annotations for 7,993 GSE127465 cells. (c) Row-normalised annotation comparison. (d) Marker expression for 12,000 Xenium cells; colours clip at \ensuremath{\pm}3, the epithelial set uses EPCAM alone, and pDC contains one cell. (e,f) Predicted cell positions and an aligned 400 \ensuremath{\mu}m H\&E field.}
\label{fig:application_celltyping}
\end{figure}
On the Xenium section, every one of the 11 predicted labels showed its highest expression on its own marker set (Fig. 3(d)). The labels were largely consistent with the unsupervised clusters (weighted purity 0.916 over 24 clusters). Mapped onto the tissue, the predictions reflected its architecture: epithelial calls overlay tumour glands, T and B calls a lymphoid aggregate, and stromal calls the intervening stroma (Fig. 3(e–f)). The Xenium comparison relies on markers rather than independent labels. The epithelial marker set contains only EPCAM, and the pDC group contains one cell.

\subsection{Molecular correspondence and whole-slide queries}
On TCGA-LUAD, the H\&E-derived profiles reproduced the patient-to-patient gradient of lymphoid genes measured in bulk RNA-seq (Fig. 4(a)). Of 317 genes, 152 correlated with bulk expression at FDR < 0.05, against a permutation-null 95th percentile of 1. The strongest included T-cell genes (CD3E r = 0.41; CD3D, CD8A, CD2). The lymphoid, smooth-muscle, stromal, endothelial and alveolar programmes were significant (r = 0.17–0.41). The epithelial/tumour programme (r = 0.09) and the myeloid programme were not (Fig. 4(b–c)).

\begin{figure}[!htbp]
\centering
\includegraphics[width=150.000mm]{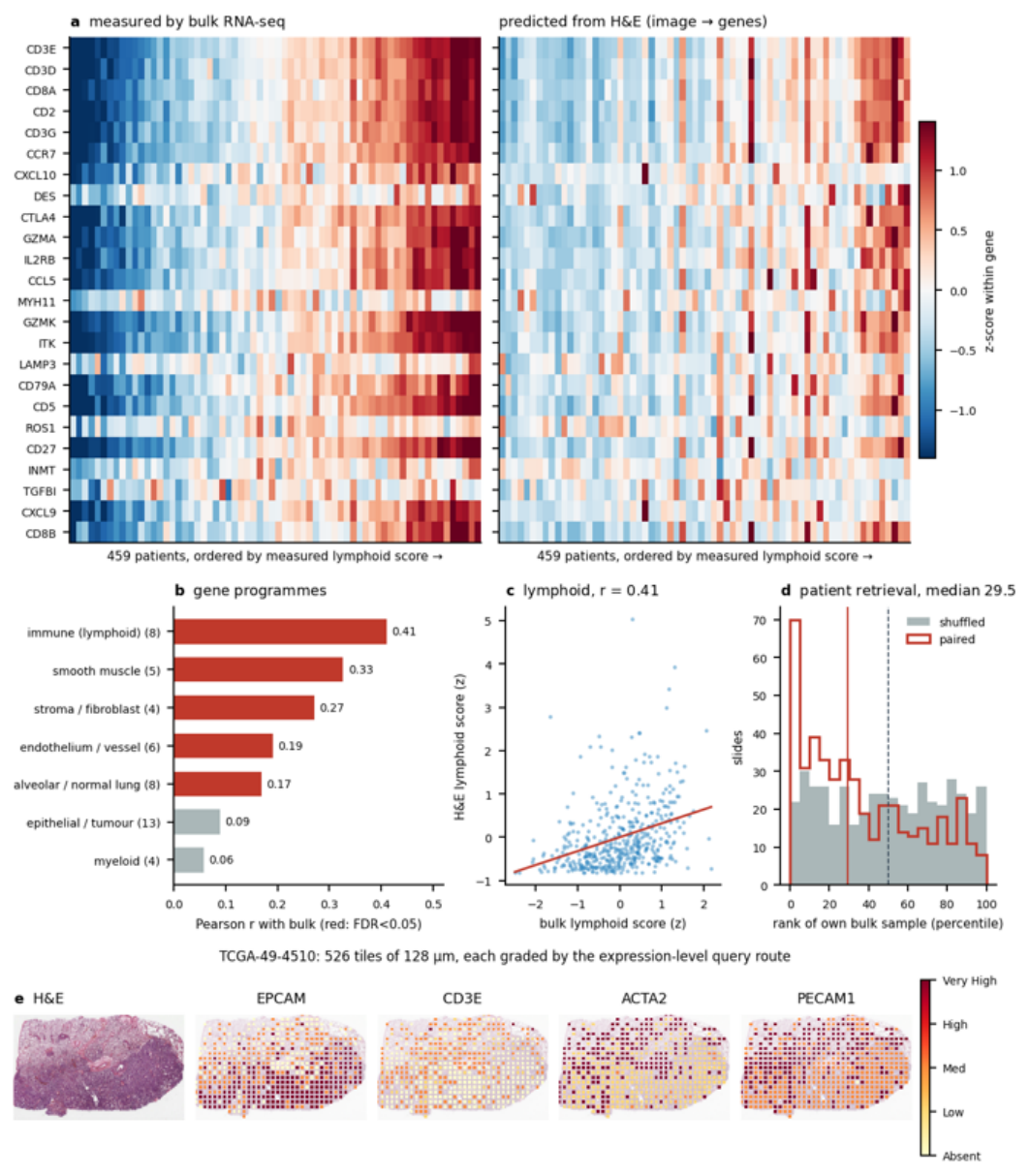}
\caption{Histology-derived molecular profiles and whole-slide queries. (a) 24 selected genes in 459 patients, ordered by measured lymphoid score and averaged over approximately seven patients per column. (b,c) Programme Pearson correlations and lymphoid association. (d) Matched-patient ranks. (e) EPCAM, CD3E, ACTA2 and PECAM1 predictions in 526 tiles of TCGA-49-4510. Queries use tile images and gene names without observed levels. Bulk RNA was extracted from a different tissue block than the diagnostic slide.}
\label{fig:application_bulk}
\end{figure}
Each slide ranked its own patient's bulk profile at a median percentile of 29.5 (chance 50), or 27.9 within its source site. The correct patient was in the top 10\% for 22\% of slides. Under shuffled pairing, the median rank returned to 48.3 and no gene was significant (Fig. 4(d)). After the technical and composition covariates, which together explained R² = 0.015, the model's profile added +0.057 in the cross-validated analysis. The strongest programme associations concern lymphoid and stromal tissue components.

As an application, we tiled one slide (TCGA-49-4510; 526 tiles of 128 \ensuremath{\mu}m) as an example. At every tile we queried the expression-level task for named genes. The displayed predictions place EPCAM in tumour-rich regions and ACTA2 and PECAM1 in stroma and vessels (Fig. 4(e)). Queries supplied each tile image and gene names without observed expression levels, outside the benchmarked Task 11 condition. Bulk RNA was extracted from a different tissue block than the diagnostic slide.

\clearpage
\section{Discussion}
VGL-stGen provides a common prediction interface for lung tissue labels, cellular identity, gene lists and queried expression levels. Its shared checkpoint connects image–expression and expression–cell-type supervision, with gene-conditioned pseudo-labels supplying image cell-type targets. The practical capability is to select a supported query according to the available image and molecular information. Region-level tissue scoring uses histology, gene-based cell typing uses ranked marker lists, and conditional expression completion also requires observed levels.

The multimodal benchmark supports the value of lung-specific adaptation under shared task protocols. Larger off-the-shelf models often score lower, and residual differences after format constraints show that output syntax alone does not explain the prediction gaps. Performance remains task-dependent: Gemma4-31B has slightly higher point estimates on the two cell-type-conditioned gene-list tasks, with paired intervals spanning zero, and VGL-stGen's tissue-classification advantage depends on pooled probability correction. Task-specific comparisons also show trade-offs: logistic regression has higher gene-classification balanced accuracy, VGL-stGen has higher macro-F1, and tumor-label performance falls within the encoder–probe range. Full-input MLPs retain higher expression pair-F1 at 75 observed levels.

Supporting experiments clarify how supervision and inputs affect the shared model. Gene-conditioned pseudo-label training improves image cell typing, and fixing supplied gene names enables level prediction from the deployed weights. Separate completion experiments show a contribution from correctly paired histology when observed expression is retained. Additional level-token supervision improves image completion, while cell-type completion shows a trade-off between pair-F1 and non-'Absent' accuracy. Those additional adapters have not been evaluated across all thirteen tasks and do not replace the deployed checkpoint in the shared-model benchmark.

The application analyses connect structured predictions to tissue-level exploration. Gene-based cell typing agrees with published annotations under the stated preprocessing and harmonised label space, and the Xenium example links predicted groups to measured marker expression. The TCGA analysis adds patient-level molecular correspondence, strongest for lymphoid and stromal programmes, in a cohort without development-patient overlap. Bulk RNA and diagnostic slides come from different tissue blocks.

The evaluation scope follows its data and inference conditions. Classification references are computationally derived and share annotation lineage with training; partial H\&E exposure and TCGA development use limit cohort independence. The masked-image expression benchmark includes samples and images represented in gene-list training. Most multimodal-comparison intervals resample items rather than sections or patients. Fixed token budgets can penalise models that generate intermediate reasoning, and native image processors and tokenizers differ. The previously analysed completion benchmark uses synthetic masking of lung marker panels. Independent annotations, verified patient separation and prospective endpoints are needed to extend the present research uses toward clinical application.

\section{Declarations}
Competing interests: Kwon Joong Na, Daeseung Lee, and Hongyoon Choi are co-founders of and hold equity in Portrai, Inc.

Code and weights: \url{https://huggingface.co/Portraier/VGL-stGen-0.5-Lung}. Gated access planned for preprint release.

\appendix
\section{Detailed methods}
\subsection{Model and adaptation}
VGL-stGen adapts the attention and feed-forward projections of the language component of \texttt{medgemma-1.5-4b-it} \citep{r01}. The vision tower is frozen. The deployed model's training stages use QLoRA with rank 16, alpha 32, dropout 0.05, a 4-bit NF4 base with double quantisation, and bfloat16 computation. Eager attention is used; gradient checkpointing is disabled in these stages. The final cancer stage uses AdamW, peak learning rate 2e-4, 500 warmup steps, weight decay 0.01, and cosine warm restarts with a minimum learning rate of 1e-6. Sequence length is 1,536 and gradient accumulation gives an effective batch of 512 from a per-device batch of one. The separate expression-adaptation experiment below starts from the assembled rank-96 adapter and uses different optimisation settings.

Training comprises gene and cell-type adaptation with stepwise prompt variants, expression bins and stain augmentation; gene-mediated image-to-cell-type distillation; joint cancer training with transcriptomic replay; and checkpoint averaging. The cancer stage trains eleven objectives: nine replayed transcriptomic tasks and two cancer tasks. The distilled image-to-cell-type objective and the three stepwise prompt variants receive no direct supervision during cancer training. The joint image-and-genes cell-type task is served without a direct training objective.

Cancer training stopped when its in-distribution validation loss plateaued. We use weight averaging \citep{r30}, representing the mean of six late adapters at rank 96 through concatenated low-rank factors and the corresponding adapter scaling. The averaging window was selected after comparing fifteen saved checkpoints on the 39-slide TCGA development subset. That subset is included in the expanded 452-slide evaluation.

The adapter was trained on a quantised base and also merged into bfloat16 weights. Matched-subset score differences reach 0.0049, with the merged path higher for cancer and the adapter path higher for cell type. The main binary tumour benchmark uses merged weights; the main nine-class tissue and cell-type benchmarks use the adapter. Packaging, prompt and scoring path are therefore part of the reported operating condition.

\subsection{Training data and gene panels}
The spatial training subset comprises 69 ST–H\&E samples in PortraiATLAS, supplemented by single-cell reference profiles from the LuCA atlas \citep{r32}. Each sample pairs spatial gene expression with histology; the sample count does not denote independent patients. The combined transcriptomic corpus contains 1,745,240 cells or spots.

A 2,000-gene panel was formed by ranking genes according to the number of source samples in which they were selected as highly variable. Gene identifiers were uppercased and punctuation other than hyphens was removed during preparation of the gene lists. The observed training top-gene lists contain 1,978 distinct identifiers, whereas the corresponding lists in the external benchmark contain 1,051. The two vocabularies share 1,049 identifiers; 929 occur only in the training lists and two only in the external lists. These counts describe observed gene-list vocabularies, not the full gene content of the assay panels. Gene lists are generated without a vocabulary constraint.

Expression tasks use Absent for undetected counts and Low, Med, High and Very High for quartiles of detected expression within a cell or Visium spot. Rows sample 100 genes across bins, with a nominal allocation of 50 Absent, 15 Low, 15 Med, 10 High and 10 Very High. Gene order is shuffled before masking. These tasks produce 2,659,430 rows before stain expansion and 6,808,162 afterwards. Observed masked-expression training data cover 2,000 genes for the image task and 1,478 for the cell-type task. These task-specific counts differ from the shared encoded MLP vocabulary described below. The levels-to-cell-type prompt exposes only the first ten sampled gene–level pairs.

The cancer corpus contains 30,965 unique image patches and 92,895 task rows. The public annotated lung tissue source associated with ARA-CNN supplies 18,554 unique patches \citep{r26}. Pathologist annotations for this project cover the same 47 Xenium samples used for spatial training and three additional public tissue sections, contributing 12,411 retained non-overlapping patches with at least 80\% labelled-region coverage. The binary manifest contains 12,411 task rows from each cancer data source. Both sources use nine tissue categories; the main binary mapping defines cancer as \texttt{TUMOR} only.

\subsection{Augmentation, sampling and stepwise instructions}
Spatial-transcriptomic and pathologist-derived patches were extracted using a target resolution of 0.5 \ensuremath{\mu}m per pixel and a target size of 224 \ensuremath{\times} 224 pixels, corresponding to a nominal 112-\ensuremath{\mu}m field of view. The LubLung source crops cover approximately 87 \ensuremath{\mu}m at 172 \ensuremath{\times} 172 pixels; these crops were resized to 221 \ensuremath{\times} 221 pixels and padded to 224 \ensuremath{\times} 224 pixels without adding tissue context. TCGA classification patches were extracted at approximately 87 \ensuremath{\mu}m, whereas the TCGA application analyses used a 128-\ensuremath{\mu}m field of view. Two offline stain-normalized appearances supplement original image-bearing transcriptomic rows. The public cancer source is also stain-expanded; multiplicity differs across sources and tasks. Validation and test manifests are unexpanded. Earlier training additionally uses on-the-fly stain augmentation, including RandStainNA \citep{r33} and percentile normalization, with geometric and photometric transformations. The distillation and final cancer runs disable on-the-fly augmentation and retain the offline variants.

The final sampler groups tasks into image-conditioned gene tasks, expression tasks, text-only gene and cell-type tasks, and cancer. Their initial shares are 0.30, 0.25, 0.20 and 0.25. Subsequent shares are computed from a softmax of the preceding epoch's family validation losses, clamped to [0.10, 0.50] and renormalised. This gives the small cancer corpus an explicit sampling share. The token loss uses focal weighting \citep{r34} with gamma 2.0 and label smoothing 0.1. Inverse-frequency sample weights apply to the nine-class tissue task.

Three auxiliary task variants use step-by-step instructions (Table 5). Their prompts ask for biological interpretation before gene-based cell typing, cell-type characteristics before gene prediction, or morphology and biological processes before image-based gene prediction. Each variant samples one fifth of its source-task rows and retains the original class-label or gene-list targets. The resulting augmentation changes the instruction, without adding intermediate reasoning targets. Across training stages, twelve task types receive target supervision; the three instruction variants bring the objective count to fifteen. The deployed interface returns the requested labels or gene lists directly.

\begin{table}[!htbp]
\centering\small
\caption{Stepwise prompt variants used in earlier training. Answer targets remain the original class labels or gene lists.}
\label{tab:templates}
\begin{tabularx}{\textwidth}{@{}>{\hsize=0.840\hsize\raggedright\arraybackslash}X>{\hsize=0.690\hsize\raggedright\arraybackslash}X>{\hsize=1.470\hsize\raggedright\arraybackslash}X@{}}
\toprule
Variant & Rows & Source \\
\midrule
Genes \ensuremath{\rightarrow} cell type & 116,278 & Single-cell atlas \\
Cell type \ensuremath{\rightarrow} genes & 115,207 & Single-cell atlas \\
Image \ensuremath{\rightarrow} genes & 1,047,500 & Spatial samples; stain-expanded \\
\bottomrule
\end{tabularx}
\end{table}
\subsection{Gene-mediated self-distillation}
Spatial data supply image–gene pairs and the atlas supplies gene–cell-type pairs. A gene-conditioned teacher scores each of the twelve cell-type labels by summing its answer-token log-probabilities. Its highest-scoring label becomes the target for the corresponding spatial image. The gap between the two highest scores defines the margin; a threshold of 1.0 retains 1.32M of approximately 1.75M candidate items. The 1.75M figure counts distillation candidates and does not denote an additional disjoint cohort.

The student uses 300,000 pseudo-labelled image rows and 200,000 replay rows for one epoch at learning rate 1e-4, with offline stain variants and eager attention. Replay retains supervision for the existing tasks during adaptation to the image-to-cell-type objective.

\subsection{Classification decoding and probability correction}
VGL-stGen closed-label generation follows a prefix tree of permitted labels. Label tokens omit leading spaces and special tokens. Under its tokenizer, first tokens are distinct within each of the twelve-class cell-type, nine-class tissue and binary label sets. Its interface can therefore return a first-token argmax and the corresponding restricted softmax score. This score is a model output, without a calibration guarantee. Comparator tokenisation and readout adaptations are specified in the multimodal benchmark protocol.

The main gene-classification comparison uses constrained generation, yielding balanced accuracy 0.6293 and macro-F1 0.6023. First-token argmax evaluated in the same run yields 0.6272 and 0.6019; the paths agree on 99.703\% of the 224,210 evaluated items. A separate generation evaluation of the same cohort yields 0.6297 and 0.6029 and provides the gene-only reference for the joint image-and-genes comparison. These are two evaluations of the same cohort.

The deployed-task evaluation and the multimodal comparison use task-specific greedy-generation limits: 8 new tokens for task 1, 16 for tasks 2–5, 160 for tasks 6–7, 56 for tasks 8–9, 128 for task 10 and 16 for task 13. Tasks 11–12 instead make 25 autoregressive expression-level decisions with fixed query names. Token limits are counted using each model's tokenizer. VGL-stGen places text before the image and uses its fast image processor (\texttt{use\_\allowbreak fast=True}); tasks 10–13 use eager attention, while tasks 1–9 use scaled dot-product attention. The multimodal benchmark preserves these task-specific inference settings while using each comparator's native chat template and image processor. The separate expression-adaptation study keeps its common autoregressive readout fixed across its trained checkpoints.

For the nine-class tissue-classification analysis, each class probability is divided by its mean across the target group, then the class scores are renormalised. The reported analysis pools first-token probabilities over the entire 39-slide subset. Macro-F1 changes from 0.1200 to 0.2801 on those same predictions. Estimation of the divisor uses no labels, but the choice to apply correction followed labelled evaluation. The analysis is therefore exploratory and transductive. Per-slide grouping defines a different operating condition.

For genes-to-cell-type, the analogous first-token correction increases balanced accuracy from 0.6272 to 0.6876 and decreases macro-F1 from 0.6019 to 0.4452. The main comparisons use uncorrected predictions. This trade-off also illustrates why a single classification metric is insufficient here.

\subsection{External cohorts and annotation dependence}
The four external spatial sections are public 10x lung cancer datasets \citep{r31}: Visium CytAssist squamous carcinoma (3,775 spots), Visium CytAssist large cell neuroendocrine carcinoma (6,189 spots), Xenium adenocarcinoma with multimodal segmentation (73,870 cells), and Xenium immuno-oncology with a custom add-on (140,376 cells). These counts are the processed evaluation units. The accompanying data-source manifest lists the official access links for these four datasets.

TACCO \citep{r35} transfers these coarse LuCA cell-type annotations to each section; the largest transferred component defines the reference class. The resulting targets share an atlas and label definition with training. The pool contains 214,246 Xenium centre-cell questions and 9,964 Visium dominant-region questions. Xenium contributes 95.6\% of the evaluated items. The pooled score therefore primarily reflects the single-cell setting.

All four sections provide measured expression and were excluded from transcriptomic training. H\&E from Visium LCNEC and the two Xenium sections supplies 4,846 of 24,822 binary cancer training rows; the Visium SCC source is absent from that manifest. This source-level check does not establish patient independence across all training stages. The two image-conditioned cell-type tasks and the two image-conditioned gene-list tasks therefore include previously seen tissue with unseen task labels. Gene-only tasks have no direct image input, but share the adapted weights. The pre-cancer cell-type checkpoint precedes this cancer-stage exposure; the deployed checkpoint follows it.

TCGA-LUAD diagnostic images \citep{r36} provide 357,692 patches from 452 slides and 398 patients, using an 87 \ensuremath{\mu}m field of view. VGL-stGen predictions are evaluated against reference labels from SegLungTCGA maps generated by ARA-CNN \citep{r26}. TCGA supplies the tissue images; ARA-CNN supplies the computational reference annotations. ARA-CNN and the cancer training corpus share LubLung provenance. Neither VGL-stGen nor the deployment probes use TCGA for gradient training, but 39 of these slides informed model development, label mapping and checkpoint selection. The expanded evaluation includes those slides. Agreement with these computational labels cannot establish independent pathologist-level validity.

The \texttt{TUMOR} class accounts for 15.09\% of evaluated patches. An always-negative classifier therefore scores 0.849 raw accuracy and 0.500 balanced accuracy. We report balanced accuracy as the principal binary metric, together with AUROC.

\subsection{Gene classifiers and adaptation-capacity control}
External comparison inputs are the same twenty gene symbols in descending expression order. Realised coverage depends on each model's vocabulary. The simple classifiers use a vocabulary learned from training lists, with either a multi-hot encoding or a rank weight \texttt{(n-r)/\allowbreak n}, where \texttt{r} starts at zero and \texttt{n} is the list length. Training lists include mixed lengths; external lists contain twenty genes. Logistic regression uses balanced class weights, the SAGA solver and up to 200 iterations. The MLP uses one hidden layer of 256 units, up to 40 iterations and internal early stopping, without class weights. Model hyperparameters are fixed.

The frozen cell-model panel contains Geneformer V2-104M and Geneformer V2-104M\_\allowbreak CLcancer \citep{r45}, and two C2S-Scale variants, with generic-language controls. Inputs preserve rank and are mapped to native tokens or gene text. This evaluates ranked marker lists, rather than each model's native full-profile task. scGPT is excluded because its expression-value input \citep{r37} would require an additional rank-to-value convention for this benchmark; its suitability for expression-valued assays is outside this comparison.

Both Geneformer variants are fine-tuned on 581,394 source rows for two epochs, batch size 256 and maximum learning rate 5e-5. Training uses class-weighted cross-entropy, AdamW with weight decay 0.01, a OneCycleLR schedule and gradient clipping at 1.0. The final-epoch weights are evaluated on 224,210 external items.

Balanced accuracy and macro-F1 are reported together: logistic regression leads on the former and VGL-stGen on the latter. The gene-classification MLP comparison is provisional because the reported fit and subsequent refits have not been reconciled. Its 0.5962 macro-F1 is shown descriptively and does not support a stable superiority claim for VGL-stGen. This comparator is distinct from the expression MLPs fitted and evaluated below.

The adaptation-capacity control compares two readouts on the same backbone, loaded in bfloat16. The generative readout retains the pretrained language-model head and trains on label-answer tokens. The discriminative readout adds a newly initialised twelve-way linear head and uses class cross-entropy. Both use rank 16, alpha 32, learning rate 1e-4, batch eight, 6,000 updates, a 300,000-row training pool and 40,000 evaluation items. The discriminative and generative arms have 32,819,212 and 32,788,480 trainable parameters, respectively. Their macro-F1 scores are 0.5209 and 0.5683. The output head and loss change jointly, so this comparison concerns their combined configuration. No repeated-seed interval is available.

\subsection{Image encoders and within-cohort reference fits}
The cell-type morphology comparison uses fifteen frozen encoders, each followed by a class-balanced logistic probe with \texttt{C=1.0}, L-BFGS and up to 2,000 iterations. Extracted encoder features receive no additional standardisation. Transferred probes train on gene-teacher pseudo-labels and are evaluated on 54,058 external items. Within-cohort references use five-fold stratified cross-validation over cells. VGL-stGen's main image-to-cell-type scores use all 224,210 items; on the same 54,058-item subset as the probes, its three evaluated checkpoints score 0.1607, 0.2268 and 0.1998.

The cancer comparison uses fourteen frozen encoders and a colour-histogram reference. Their deployment probes train on the same 30,965 unique cancer patches as VGL-stGen. Cancer features are L2-normalised; class-balanced logistic regression selects regularisation by grouped cross-validation within the source training data. A binary probe supplies the main comparison; nine-class-derived binary results are separate diagnostic arms. The model's own medically adapted vision tower, called SigLIP in the result files, is also fitted with a linear probe as an additional control.

For cancer reference fits, the target cohort is split into three patient-grouped folds. Each held-out patch receives an out-of-fold prediction. A matched-size arm trains on 30,965 target-fold patches at the source class balance; a full arm uses all available training-fold patches. Both use the regularisation selected on the matched subset. The full arm is not separately tuned, and the direction of any tuning effect is unknown. These comparisons constrain a simple sample-count explanation but do not isolate a causal contribution of cohort shift.

The gene reference fits use five-fold cell-level cross-validation on approximately 100,000 external items. Cell and section dependence can make these references optimistic. The cancer folds instead separate patients. Scores from these different constructions should not be interpreted as directly comparable limits of biological information.

Within-cohort probe scores provide empirical supervised references for each representation. The transfer plots place target-cohort probe balanced accuracy on the horizontal axis and transferred balanced accuracy on the vertical axis. Both axes show empirical scores, not mathematical performance bounds. For imaging, VGL-stGen is compared with the reference from a probe on its own frozen tower; VGL-stGen itself is not adapted within the target cohort.

\subsection{Gene-list evaluation}
Four gene-list tasks use a seeded sample of 1,000 external items. Generation targets twenty genes; completion masks five of the twenty and supplies the remaining fifteen. The sample contains 949 Xenium cells and 51 Visium spots, with three cell types represented by at most three items. Gene-set F1 is \texttt{2|A \ensuremath{\cap} B| /\allowbreak  (|A| +\allowbreak  |B|)}, averaged over items. These samples provide task-level estimates with little support for subgroup claims.

Completion evaluation prompts specify the actual number of withheld genes. Some training prompts specified output counts that differed from their targets.

Training-global frequency rules provide transferable references. External-global and external-cell-type frequency rules use target-cohort statistics and provide descriptive comparisons with additional information. On image-to-gene generation, the training-global rule scores 0.0367, the external-global rule 0.2913 and the external-cell-type rule 0.4009. The last comparison changes both cohort and cell-type conditioning. None of these scores is a mathematical upper bound or a causal decomposition of the transfer gap.

The frequency analysis counts one row per cell or spot: 1,221,886 training items and 224,210 external items, compared over 1,049 shared genes. The top-twenty frequency lists share five symbols. Panel coverage, cell composition and assay processing all differ between the cohorts, so their separate effects remain unresolved.

Internal gene-list F1 is 0.4192, 0.3943, 0.3310 and 0.3972 for the four tasks. Within the image-to-gene split, recorded test cell/spot identifiers do not overlap its training split, but all 69 internal image-test samples also occur in training. The internal cell-type-to-gene evaluation covers effectively 4 classes. These conditions differ from the expression-task split described below.

For a partial gene list without an image or cell-type input, the interface composes cell-type prediction with cell-type-conditioned completion. Its recorded F1 is 0.0472, compared with 0.0484 when supplied the reference cell type; the intermediate cell-type prediction is correct on 81.6\% of sampled items. This is a composed interface operation rather than an additional directly trained task.

\subsection{Expression evaluation and image-shuffle controls}
The expression-task training and test manifests have disjoint sample and recorded cell/spot identifiers. Each task is scored on 1,000 existing benchmark rows. The masked-image evaluation contains 665 Xenium cells and 335 Visium spots from seven samples. All seven samples are represented in gene-list training, and 692 benchmark cells or spots have matching identifiers and identical source images in that training set. The masked-cell-type benchmark uses the atlas-level identifier ‘LuCA\_\allowbreak NCSCLC’, which is also present in gene-list training; its generated row-based cell identifiers do not establish a cross-task cell-barcode or donor linkage. These internal benchmarks therefore do not establish patient-independent generalisation across the full training sequence. The benchmark had already been analysed before the controlled adaptation study.

In the deployed model, the masked-expression objectives were trained without conditioning gene lists. Both decoding procedures use the same checkpoint and evaluation rows. This comparison measures inference behaviour for those weights. The separate adaptation experiment supplies fully conditioned examples to both loss conditions and holds their inference procedure fixed.

Free generation is scored as emitted. Fixed-name decoding parses the twenty-five masked entries from the supplied hundred-gene list, then predicts each level from five permitted first tokens. Two independent extractors identify the same masked entries on all 176,008 held-out masked-task rows. Autoregressive decoding conditions each slot on the model's previous level answers. Pair-F1 over the fixed names equals slot-level accuracy. These results apply to the specified input format.

Per-gene and gene-by-cell-type tables are estimated from the training split. Their scores are 0.6198 and 0.6963. Both omit the seventy-five visible gene levels available to VGL-stGen. The deployed autoregressive scores are 0.6596 and 0.6858, with paired model-minus-table row intervals of [+0.0350, +0.0445] and [\ensuremath{-}0.0160, \ensuremath{-}0.0052]. Context-aware MLPs on these same rows give three-seed mean pair-F1s of 0.6811 and 0.7300, respectively. The MLPs align available inputs and targets but use different encoders and training histories.

Image-shuffle diagnostics retain the expression prompt and replace the paired image. Under teacher-forced preceding answers, the intact-minus-shuffled effect is 0.0097 [0.0071, 0.0124] for VGL-stGen and 0.0001 [\ensuremath{-}0.0010, +0.0012] for the base model. With preceding answer positions fixed to Absent, VGL-stGen's effect is 0.0092 [0.0067, 0.0119]. These settings differ from autoregressive generation and cannot be subtracted from its model-minus-table margin to assign a fraction of the gain to images.

Very-high-gene prediction is evaluated with the true answer length supplied. The model scores 0.2828; the 0.2247 frequency and 0.0144 random references use evaluation-shard gene support and answer lengths. Levels-to-cell-type prediction scores 0.7420 raw accuracy from the first ten sampled pairs. Its majority reference of 0.362 is selected within evaluation shards. These task-specific references use evaluation information, whereas the masked-completion frequency tables use training data. Reported scores aggregate four evaluation shards.

\subsection{Context-aware expression baselines}
MLPs encode all 100 supplied gene–state pairs, including 75 observed levels and 25 masks, by summed embeddings divided by ten and layer normalisation. After concatenating the extra inputs, two hidden linear layers with GELU activations and dropout 0.1 after the first hidden layer predict five levels per gene. All variants share a training-derived vocabulary of 2,000 genes plus an unknown entry; five-class cross-entropy is applied only to the 25 queried genes. The context-only image-task model receives an indicator specifying a centre-cell or tissue-region query. The full image model additionally receives frozen 1,536-dimensional UNI2-h CLS features \citep{r24}; the cell-type model receives the supplied twelve-class one-hot label plus an unknown category.

UNI2-h processes the same RGB images resized to 224 pixels and normalised by ImageNet statistics. Features are standardised using training-set means and population standard deviations, floored at 1e-4. This extractor applies no additional stain or geometric augmentation. The raw images, visible expression, query names, target levels and query-region or cell-type inputs are aligned with VGL-stGen. The comparison does not match vision encoders or staged training histories.

The original task splits supply 1,037,183 image training rows, 292,532 cell-type training rows and 10,000 validation rows per task. AdamW uses batch 1,024, weight decay 1e-4, a maximum of twenty epochs and validation patience three. A seed-42 search over widths 256/512 and learning rates 1e-3/3e-4 selects width 256 for the context-only image-task and cell-type models, and width 512 for the full image model, all at learning rate 1e-3. The selected settings are repeated at seeds 7 and 2026; each run uses its validation-selected epoch. Configurations and weights are fixed before test scoring. Image validation contains Xenium only, whereas the benchmark includes both platforms.

The context-only image-task MLP has mean pair-F1 0.6666. The full image MLP reaches 0.6811, but the two selected configurations differ in width as well as input. A separate width-512 matched-setting comparison has image-added differences of \ensuremath{-}0.0024, +0.0231 and +0.0141 across seeds; the direction is not consistent. Neither comparison establishes that images are dispensable or isolates a universal imaging advantage.

\subsection{Controlled expression adaptation and image permutation}
Both conditions initialise from the same deployed rank-96 adapter. They share 4,096 additional training rows: 1,024 Xenium image, 1,024 Visium image and 2,048 cell-type rows. Additional training uses a fixed subset of the existing task-training data. The complete 100-gene list, 75 observed levels, image or supplied cell type, answer order and row order are identical between conditions. Training uses seeds 42, 7 and 2026, with no replay of other tasks.

Full-answer supervision applies mean full-vocabulary cross-entropy to all answer tokens and end-of-turn. Level-only supervision applies it to exactly the first tokens of the 25 requested levels; other answer positions receive no direct loss. This changes supervised positions without introducing a five-class-normalised training loss or a new prediction head. Both conditions use autoregressive inference from the model's own preceding level predictions, restricted to the same five permitted first tokens.

Each run uses 256 updates, effective batch 16, AdamW learning rate 5e-5, sixteen linear warmup updates followed by a constant rate, weight decay 0.01, betas (0.9, 0.999), epsilon 1e-8 and gradient clipping at 1.0. The base is NF4 double-quantised with bfloat16 autocast; trainable language-side adapter parameters are float32, rank 96, alpha 32 and dropout 0.05. The vision tower and projector remain frozen. Gradient checkpointing is enabled. The terminal checkpoint is used without a learning-rate or intermediate-checkpoint search.

Development evaluation uses 96 Xenium, 96 Visium and 128 cell-type rows. Three Visium samples are selected by a fixed hash of sample names, excluded entirely from additional training, and sampled at 32 rows each. Those samples were present in the initial model's training corpus. Other cross-family exposure is unverified. The equal-weight mean pair-F1 across these three groups and three seeds selects level-only as the representative supervision condition. The representative seed is fixed at 42, independent of its benchmark score. Both conditions at all three seeds are retained for the reported benchmark comparison, with weights and evaluation rows fixed before scoring.

Table 6 reports level macro-F1 and accuracy on non-Absent targets; these reveal performance obscured by the frequent Absent class. Table 7 reports every seed and platform. Level-only image pair-F1 exceeds full-answer training at all seeds on both platforms. Cell-type pair-F1 does not improve beyond full-answer training, although its secondary metric means are slightly higher. The MLP retains higher means on both secondary metrics in both tasks.

\begin{table}[!htbp]
\centering\small
\caption{Secondary expression metrics on the same benchmark rows. Mean \ensuremath{\pm} sample SD across 3 individual seeds; the deployed checkpoint is a single fitted model. Non-Absent accuracy restricts targets to the 4 detected-expression levels.}
\label{tab:expression_secondary}
\begin{tabularx}{\textwidth}{@{}>{\hsize=0.600\hsize\raggedright\arraybackslash}X>{\hsize=1.080\hsize\raggedright\arraybackslash}X>{\hsize=1.160\hsize\raggedright\arraybackslash}X>{\hsize=1.160\hsize\raggedright\arraybackslash}X@{}}
\toprule
Task & Model & Level macro-F1 & Non-Absent accuracy \\
\midrule
Image & Deployed & 0.3780 & 0.2276 \\
Image & Full-answer & 0.4269 \ensuremath{\pm} 0.0024 & 0.2664 \ensuremath{\pm} 0.0038 \\
Image & Level-only & 0.4463 \ensuremath{\pm} 0.0015 & 0.2930 \ensuremath{\pm} 0.0021 \\
Image & Full-input MLP & 0.4586 \ensuremath{\pm} 0.0021 & 0.3160 \ensuremath{\pm} 0.0011 \\
Cell type & Deployed & 0.4625 & 0.3611 \\
Cell type & Full-answer & 0.5445 \ensuremath{\pm} 0.0008 & 0.4433 \ensuremath{\pm} 0.0049 \\
Cell type & Level-only & 0.5472 \ensuremath{\pm} 0.0022 & 0.4535 \ensuremath{\pm} 0.0023 \\
Cell type & Full-input MLP & 0.5675 \ensuremath{\pm} 0.0026 & 0.4786 \ensuremath{\pm} 0.0047 \\
\bottomrule
\end{tabularx}
\end{table}
\begin{table}[!htbp]
\centering\small
\caption{Every additional-training seed and image platform. The pooled image set contains 665 Xenium cells and 335 Visium spots; the cell-type set contains 1,000 rows. Difference is level-only minus full-answer pair-F1. These repeat additional training from one initial checkpoint.}
\label{tab:expression_seeds}
\begin{tabularx}{\textwidth}{@{}>{\hsize=1.200\hsize\raggedright\arraybackslash}X>{\hsize=0.500\hsize\raggedright\arraybackslash}X>{\hsize=1.100\hsize\raggedright\arraybackslash}X>{\hsize=1.100\hsize\raggedright\arraybackslash}X>{\hsize=1.100\hsize\raggedright\arraybackslash}X@{}}
\toprule
Evaluation & Seed & Full-answer & Level-only & Difference \\
\midrule
Image pooled & 42 & 0.67096 & 0.68032 & +0.00936 \\
Image pooled & 7 & 0.67252 & 0.68164 & +0.00912 \\
Image pooled & 2026 & 0.67252 & 0.67948 & +0.00696 \\
Xenium & 42 & 0.71717 & 0.72523 & +0.00806 \\
Xenium & 7 & 0.71705 & 0.72493 & +0.00788 \\
Xenium & 2026 & 0.71735 & 0.72343 & +0.00608 \\
Visium & 42 & 0.57922 & 0.59116 & +0.01194 \\
Visium & 7 & 0.58412 & 0.59570 & +0.01158 \\
Visium & 2026 & 0.58352 & 0.59224 & +0.00872 \\
Cell type & 42 & 0.72260 & 0.72028 & -0.00232 \\
Cell type & 7 & 0.72240 & 0.71964 & -0.00276 \\
Cell type & 2026 & 0.72132 & 0.72040 & -0.00092 \\
\bottomrule
\end{tabularx}
\end{table}
For all six additional-training checkpoints, rows within each sample and original platform are deterministically ordered by a SHA-256 hash of their identifiers. Each row receives the next row's image, wrapping the final row to the first. Every donor image differs from the original. Visible expression, targets and autoregressive decoding are unchanged. On the pooled image benchmark, the intact-minus-permuted pair-F1 difference is positive in each seed: full-answer 0.00504/0.00632/0.00656 and level-only 0.00796/0.00712/0.00820 for seeds 42/7/2026. Mean differences are 0.00597 and 0.00776, with sample CIs [0.00316, 0.00764] and [0.00263, 0.01042]. The paired difference between these effects is +0.00179, row CI [\ensuremath{-}0.00097, 0.00460] and sample CI [\ensuremath{-}0.00067, 0.00336]. Level-only's normal-input advantage of 0.00848 remains 0.00669 after permutation. These comparisons establish sensitivity to correctly paired images on the pooled benchmark under both objectives; they do not establish that level-only training increased that sensitivity. No full thirteen-task evaluation was performed for the additional-training checkpoints; the task inventory and classification results refer to the deployed model.

Table 8 reports every intact and permuted score in the six-checkpoint comparison.

\begin{table}[!htbp]
\centering\small
\caption{Image pairing under both supervision conditions. Each seed is evaluated on the same 1,000 rows and 25 queried levels per row with 75 observed levels. The within-sample/platform image permutation is identical across all 6 checkpoints. Positive differences indicate higher pair-F1 with intact images.}
\label{tab:expression_factorial}
\begin{tabularx}{\textwidth}{@{}>{\hsize=1.200\hsize\raggedright\arraybackslash}X>{\hsize=0.500\hsize\raggedright\arraybackslash}X>{\hsize=1.100\hsize\raggedright\arraybackslash}X>{\hsize=1.100\hsize\raggedright\arraybackslash}X>{\hsize=1.100\hsize\raggedright\arraybackslash}X@{}}
\toprule
Supervision & Seed & Intact & Permuted & Difference \\
\midrule
Full-answer & 42 & 0.67096 & 0.66592 & +0.00504 \\
Full-answer & 7 & 0.67252 & 0.66620 & +0.00632 \\
Full-answer & 2026 & 0.67252 & 0.66596 & +0.00656 \\
Level-only & 42 & 0.68032 & 0.67236 & +0.00796 \\
Level-only & 7 & 0.68164 & 0.67452 & +0.00712 \\
Level-only & 2026 & 0.67948 & 0.67128 & +0.00820 \\
\bottomrule
\end{tabularx}
\end{table}
\subsection{Completion with fewer observed expression levels}
We retain the same 1,000 image rows, 25 queried genes, target levels and image permutation while reducing observed gene levels from 75 to 50 and 25. Within each row, observed genes are deterministically ordered by a SHA-256 hash of the row identifier and upper-case gene name. The first 50 or 25 are retained in their original relative order, yielding nested subsets independent of target levels or predictions. Exact hashing specifications for gene selection and image permutation accompany the source data. Excluded observed pairs are removed from the prompt; the 25 original query masks remain. Expression-level definitions and targets are unchanged.

The VGL-stGen checkpoint is the development-selected level-only seed 42. Image-inclusive and context-only MLPs use their existing matched-width fits (512, learning rate 0.001, seeds 42/7/2026). Excluded MLP input tokens become padding; query masks, the recipient's centre-cell/region query indicator, training-derived feature normalisation and all weights remain fixed. Only image features are permuted in the image-inclusive MLP. These are inference-time stress conditions for models trained with 75 observed levels, with no budget-specific retraining.

Table 9 gives all budgets, image conditions and secondary metrics. The primary budget contrast is the intact-minus-permuted image effect at 25 observed levels minus that at 75. Paired row and sample bootstrap preserve the original platform mixture, conditional on the fixed models, one image permutation and one nested-subset rule. This analysis evaluates reduced input context within synthetic marker-panel queries; it does not establish performance for a smaller assay panel or measurement-cost savings.

\begin{table}[!htbp]
\centering\small
\caption{Synthetic completion under reduced observed context, with the same 25 queries. Obs. denotes observed gene levels. VGL-stGen uses the development-selected level-only seed 42; MLPs report mean \ensuremath{\pm} sample SD across 3 fixed seeds (width 512, learning rate 0.001). All models retain weights trained with 75 observed levels. The context-only MLP has no image input.}
\label{tab:expression_budget}
\begin{tabularx}{\textwidth}{@{}>{\hsize=0.400\hsize\raggedright\arraybackslash}X>{\hsize=1.450\hsize\raggedright\arraybackslash}X>{\hsize=1.050\hsize\raggedright\arraybackslash}X>{\hsize=1.050\hsize\raggedright\arraybackslash}X>{\hsize=1.050\hsize\raggedright\arraybackslash}X@{}}
\toprule
Obs. & Model / image & Pair-F1 & Level F1m & Non-Absent acc. \\
\midrule
75 & VGL-stGen / Intact & 0.6803 & 0.4466 & 0.2930 \\
75 & VGL-stGen / Permuted & 0.6724 & 0.4308 & 0.2727 \\
75 & Image MLP / Intact & 0.6811 \ensuremath{\pm} 0.0038 & 0.4586 \ensuremath{\pm} 0.0021 & 0.3160 \ensuremath{\pm} 0.0011 \\
75 & Image MLP / Permuted & 0.6700 \ensuremath{\pm} 0.0030 & 0.4375 \ensuremath{\pm} 0.0005 & 0.2903 \ensuremath{\pm} 0.0026 \\
75 & Context MLP / — & 0.6695 \ensuremath{\pm} 0.0092 & 0.4524 \ensuremath{\pm} 0.0086 & 0.3187 \ensuremath{\pm} 0.0055 \\
50 & VGL-stGen / Intact & 0.6681 & 0.4173 & 0.2593 \\
50 & VGL-stGen / Permuted & 0.6609 & 0.4014 & 0.2401 \\
50 & Image MLP / Intact & 0.6674 \ensuremath{\pm} 0.0048 & 0.4437 \ensuremath{\pm} 0.0019 & 0.3096 \ensuremath{\pm} 0.0016 \\
50 & Image MLP / Permuted & 0.6557 \ensuremath{\pm} 0.0054 & 0.4231 \ensuremath{\pm} 0.0036 & 0.2854 \ensuremath{\pm} 0.0003 \\
50 & Context MLP / — & 0.6477 \ensuremath{\pm} 0.0116 & 0.4278 \ensuremath{\pm} 0.0080 & 0.3038 \ensuremath{\pm} 0.0040 \\
25 & VGL-stGen / Intact & 0.6599 & 0.3839 & 0.2164 \\
25 & VGL-stGen / Permuted & 0.6496 & 0.3631 & 0.1928 \\
25 & Image MLP / Intact & 0.6469 \ensuremath{\pm} 0.0054 & 0.4215 \ensuremath{\pm} 0.0031 & 0.2968 \ensuremath{\pm} 0.0040 \\
25 & Image MLP / Permuted & 0.6330 \ensuremath{\pm} 0.0078 & 0.3984 \ensuremath{\pm} 0.0059 & 0.2705 \ensuremath{\pm} 0.0015 \\
25 & Context MLP / — & 0.6072 \ensuremath{\pm} 0.0152 & 0.3903 \ensuremath{\pm} 0.0079 & 0.2805 \ensuremath{\pm} 0.0025 \\
\bottomrule
\end{tabularx}
\end{table}
For the fixed VGL-stGen checkpoint, intact-image pair-F1 is 0.68032, 0.66808 and 0.65992 at 75, 50 and 25 observed levels. Image-pairing effects are +0.00796, +0.00720 and +0.01032; the latter two have sample CIs [0.00200, 0.01125] and [0.00561, 0.01443]. The 25-minus-75 change in image effect is +0.00236, row CI [\ensuremath{-}0.00180, 0.00644] and sample CI [\ensuremath{-}0.00037, 0.00543], leaving a pooled increase unresolved. The corresponding point changes differ across Xenium (+0.00481) and Visium (\ensuremath{-}0.00251).

At 25 observed levels, VGL-stGen pair-F1 is higher than the image-MLP mean (0.65992 versus 0.64691), while non-Absent accuracy (0.2164 versus 0.2968) and level macro-F1 (0.3839 versus 0.4215) are lower. These results document a metric-specific trade-off under reduced context, rather than an overall completion advantage. All budgets and image conditions are retained in the table and accompanying source data.

\subsection{Alignment checks and cancer label definitions}
Segmentation colour maps are aligned to slides by axis-wise linear scaling. A tissue-mask comparison searches translations over \ensuremath{\pm}8\% of a common grid; the unshifted alignment is optimal on all 452 slides, aspect-ratio error is at most 0.0009, and horizontal or vertical flips do not improve the fit. This evaluates geometric registration independently of the biological correctness of the segmentation labels.

Tissue-mask quality control identifies 4,999 nearly empty tiles, or 1.40\% of the evaluated set. Excluding them changes balanced accuracy from 0.7532 to 0.7519; restricting evaluation by the alignment threshold gives 0.7596. These post-hoc sensitivity analyses accompany the unfiltered estimate.

To assess overlap with known development patients, we exclude all slides belonging to the 39 patients represented by the 39-slide development subset. This removes 40 slides and 31,574 patches, leaving 412 slides, 359 patients and 326,118 patches. With unchanged predictions, VGL-stGen's observed balanced accuracy is 0.7523, patient-bootstrap CI [0.7407, 0.7628], and AUROC 0.8253, compared with observed balanced accuracy 0.7534 on the full cohort. The original 0.7532 headline is a slide-bootstrap mean. Comparator probes are reconstructed using their original source-training settings and verified against their full-cohort scores before the patient exclusion. VGL-stGen remains sixth among the fifteen systems; the own-tower probe is an additional control outside this ranking. This sensitivity analysis does not remove adaptive development, shared annotation lineage or possible foundation-model pretraining exposure.

The label-definition comparison refits each encoder probe under the wide mapping (\texttt{TUMOR}, \texttt{NECROSIS} and \texttt{MIXED} positive) and the \texttt{TUMOR}-only mapping on the same 452-slide cohort. Its median paired balanced-accuracy difference is 0.1191 across the fourteen encoders and 0.0436 for the colour reference. Each encoder's patient-paired interval excludes zero. Both target definition and fitted classifier change, so the result measures benchmark sensitivity rather than the clinical superiority of one mapping. The mapping was chosen during development.

\subsection{Statistical analysis}
Binary cancer intervals use 1,000 slide-bootstrap draws, with patient-bootstrap intervals to account for patients contributing multiple slides. The empirical balanced accuracy is 0.7534 and its bootstrap mean is 0.7532. The slide interval is [0.7452, 0.7610]; the patient interval is [0.7441, 0.7622]. Neither incorporates annotation error, repeated-training variation or selection bias.

The principal paired cell-type comparisons use 2,000 hierarchical bootstrap draws, resampling four sections and then items within each sampled section. The same resampled indices are used for both systems. With only four sections, these intervals remain sensitive to the observed cohort composition. They should be read alongside section-specific results, not as strong evidence about a population of new patients. The retained Geneformer fine-tuning comparison instead uses cell-level resampling. The deployed model's expression and image-shuffle intervals resample rows.

The controlled adaptation study reports means and sample standard deviations across three individually fitted models, not ensembles. Paired contrasts first average matched-seed per-row differences and then use 2,000 bootstrap draws. Image row intervals resample rows within platform; image sample intervals resample samples with replacement within platform, retain all their rows and preserve the fixed Xenium/Visium mixture. Sample identifiers are not verified patients. The cell-type benchmark has only one sample identifier, so no sample-generalisation interval is estimated. Intervals condition on the fitted models and do not estimate a population of training seeds. The three seeds vary additional training from a common initial checkpoint, not the entire pretraining and adaptation history.

Balanced accuracy and macro-F1 are reported jointly. Confidence intervals for model-performance comparisons are not adjusted for multiple comparisons. A confidence interval crossing zero is not evidence of equivalence or non-inferiority.

\subsection{Data and code availability}
Code and weights are hosted at \url{https://huggingface.co/Portraier/VGL-stGen-0.5-Lung}, with gated access planned for the preprint release. Reproduction of the evaluations also requires the processed manifests, annotation-transfer outputs and comparison predictions; internal expression results require the corresponding held-out split.

\section{TCGA bulk RNA-seq analysis}
\subsection{Gene profiles and molecular endpoints}
The analysis includes 459 TCGA-LUAD patients with readable diagnostic FFPE slides and primary-tumour bulk RNA-seq \citep{r36}. The deployed checkpoint predicted twenty gene symbols for each of 112 random tissue patches per slide at a 128 \ensuremath{\mu}m field of view. Each slide's gene score is its fraction of patches naming that gene. The 317 evaluated genes satisfy both eligibility criteria: occurrence in predictions for at least 10\% of slides and measured expression above one TPM in at least half of patients.

Measured expression was transformed as log2(TPM+1), and both measured and image-derived matrices were centred within gene. This centres the analysis on between-patient variation for each gene. Per-gene Pearson correlations were assessed with BH-FDR, with shuffled patient pairing providing a null comparison. Marker-programme scores were compared separately. Lymphoid, smooth-muscle, stromal, endothelial and alveolar programmes met FDR < 0.05; epithelial/tumour and myeloid programmes did not. Proliferation was not scored because MKI67 was the only eligible proliferation marker.

Fig. 4(b) shows the programme correlations and the significance marked in the original plot. Of 317 genes, 152 met the FDR threshold, versus a permutation-null 95th percentile of one.

\subsection{Patient matching and covariate controls}
Each slide's matching bulk profile was ranked among all 459 patients. The median rank was the 29.5th percentile, compared with chance 50; restricting candidates to the same tissue source site gave 27.9. The correct patient was in the top 10\% for 22\% of slides. Shuffling slide–bulk pairing gave a median rank of 48.3 and no significant genes. Sequencing depth, magnification, source site and tumour fraction together explained cross-validated R² = 0.015; adding the image-derived gene profile increased R² by 0.057.

\subsection{Visualisation and scope}
Fig. 4(a) displays 24 selected genes in 459 patients, ordered by measured lymphoid score and averaged over approximately seven adjacent patients per displayed column. Both heatmaps are z-scored within gene. Programme associations and the paired versus shuffled patient-rank distributions appear in panels (b–d).

The bulk measurements and diagnostic slides come from different tissue blocks. The separate named-gene maps in Fig. 4(e) illustrate applying molecular queries across one slide and are described in the application methods.

\section{Application analysis methods}
\subsection{Single-cell and Xenium cell typing}
We evaluated genes \ensuremath{\rightarrow} cell type (Task 1) with the checkpoint (merged bfloat16 weights, constrained decoding). GSE127465 comprises seven treatment-naive NSCLC patients with published annotations \citep{r46}. After removal of housekeeping, mitochondrial and ribosomal genes, the analysis retains 2,000 highly variable genes and samples 7,993 tumour-tissue cells. Each input contains the twenty highest-expressed retained genes in rank order. Macrophage/monocyte and conventional dendritic-cell predictions are combined as MoMacDC to match the eleven-class reference. Patient-specific malignant clusters are treated as epithelial, with an additional analysis excluding them.

The 10x Xenium Human Lung Cancer FFPE application \citep{r31} samples 12,000 cells with at least ten transcripts from a 377-gene panel. Marker scores z-score each gene across cells and average within canonical sets; 39 of 78 markers are present on the panel. Fig. 3 shows the GSE127465 comparison, the Xenium marker matrix and an aligned post-Xenium H\&E field. The epithelial marker set contains EPCAM alone and the predicted pDC group contains one cell. The Xenium comparison relies on markers rather than independent labels.

Balanced accuracy is 0.815 and macro-F1 0.819, with balanced accuracy 0.814 after malignant-cell exclusion and 0.70–0.86 per patient. Major confusions were that 47\% of NK cells were called T cells, 46\% of pDCs were called plasma cells, and 22\% of B cells were called T cells. Confidence was lower for incorrect calls (median 0.47 vs 0.83). Balanced accuracy falls to 0.775 after gene-order shuffling and 0.282 without filtering. Xenium weighted purity is 0.916 across twenty-four unsupervised clusters. These application results use their own preprocessing and harmonised label space and are separate from the main 224,210-item evaluation.

\subsection{Whole-slide named-gene illustration}
Fig. 4 uses TCGA-49-4510 with 526 tiles of 128 \ensuremath{\mu}m. Each tile image was paired with gene names and no observed expression levels, outside the benchmarked Task 11 input condition. Four displayed maps concern EPCAM, CD3E, ACTA2 and PECAM1. High was not emitted in this application; four of the five permitted levels were used. Bulk RNA was extracted from a different tissue block than the diagnostic slide.

\section{Multimodal benchmark protocol}
The benchmark evaluates VGL-stGen and ten off-the-shelf medical or general-purpose multimodal models on the same thirteen tasks. The comparator families are MedGemma \citep{r01}, Gemma 3 \citep{r47}, Gemma 4 \citep{r48}, Qwen3-VL \citep{r49}, Qwen3.8 \citep{r50}, DeepSeek-VL2 \citep{r51} and GLM-4.6V-Flash \citep{r52}. The full GLM-4.6V checkpoint is not included. Table 10 specifies the task-level settings. Task identifiers refer to the deployed task inventory.

\begin{table}[!htbp]
\centering\small
\caption{Fixed task evaluation settings shared across models. Tokens are maximum generated tokens; T11–12 instead make 25 autoregressive level choices per row. T2–3 preserve the original mixed-task batch composition. Counts are per task. All tasks except T4 use NF4; T4 uses BF16. Task key: T1, genes \ensuremath{\rightarrow} cell type; T2, image + genes \ensuremath{\rightarrow} cell type; T3, image \ensuremath{\rightarrow} cell type; T4, image \ensuremath{\rightarrow} binary tumor status; T5, image \ensuremath{\rightarrow} nine-class tissue type; T6, image \ensuremath{\rightarrow} gene names; T7, cell type \ensuremath{\rightarrow} gene names; T8, image + partial gene list \ensuremath{\rightarrow} missing gene names; T9, cell type + partial gene list \ensuremath{\rightarrow} missing gene names; T10, image \ensuremath{\rightarrow} very-high-expression gene names; T11, image + observed levels \ensuremath{\rightarrow} queried levels; T12, cell type + observed levels \ensuremath{\rightarrow} queried levels; T13, gene–level pairs \ensuremath{\rightarrow} cell type.}
\label{tab:vlm_protocol}
\begin{tabularx}{\textwidth}{@{}>{\hsize=0.400\hsize\raggedright\arraybackslash}X>{\hsize=1.480\hsize\raggedright\arraybackslash}X>{\hsize=0.920\hsize\raggedright\arraybackslash}X>{\hsize=1.200\hsize\raggedright\arraybackslash}X@{}}
\toprule
Tasks & Items per task & Output limit & Batch / shards \\
\midrule
1 & 224,210 cells/spots & 8 tokens & 16 / 4 \\
2–3 & 224,210 cells/spots & 16 tokens & 16 mixed / 4 \\
4 & 357,692 patches / 452 slides & 16 tokens & 32 / 8 slide shards \\
5 & 30,719 patches / 39 slides & 16 tokens & 32 \\
6–7 & 1,000 rows & 160 tokens & 8 / 4 \\
8–9 & 1,000 rows & 56 tokens & 8 / 4 \\
10 & 1,000 rows & 128 tokens & 8 / 4 \\
11–12 & 1,000 rows / 25,000 levels & 25 level choices & 1 / 4 \\
13 & 1,000 rows & 16 tokens & 8 / 4 \\
\bottomrule
\end{tabularx}
\end{table}
The original task pipeline supplies inputs and instructions, collates batches, extracts answers and computes scores. The same canonical image files and molecular context are supplied to every model, with model-native image processing and chat templates. Consequently, matched inputs do not imply identical resized tensors, visual-token counts or tokenized instruction lengths. Table 11 identifies the exact model repositories and image-processing configurations. Full revisions, library versions and source hashes are included with the ancillary benchmark data.

\begin{table}[!htbp]
\centering\small
\caption{Model repositories and native image processing. VGL-stGen additionally uses the deployed lung adapter (merged for binary tumor classification, T4). Full revisions, processor settings, library versions and loading exceptions are supplied in the ancillary model-conditions files. Budgets describe configuration, not measured per-image token counts.}
\label{tab:vlm_models}
\begin{tabularx}{\textwidth}{@{}>{\hsize=0.600\hsize\raggedright\arraybackslash}X>{\hsize=0.960\hsize\raggedright\arraybackslash}X>{\hsize=1.440\hsize\raggedright\arraybackslash}X@{}}
\toprule
Model & Repository & Image processing \\
\midrule
VGL-stGen & google/medgemma-1.5-4b-it & 896 \ensuremath{\times} 896; mean/std 0.5 \\
MedGemma-1.5-4B & google/medgemma-1.5-4b-it & 896 \ensuremath{\times} 896; mean/std 0.5 \\
Gemma3-4B & google/gemma-3-4b-it & 896 \ensuremath{\times} 896; mean/std 0.5 \\
Qwen3-VL-8B & Qwen/Qwen3-VL-8B-Instruct & Dynamic resolution; 16-pixel patches; merge 2 \\
Qwen3-VL-32B & Qwen/Qwen3-VL-32B-Instruct & Dynamic resolution; 16-pixel patches; merge 2 \\
Gemma4-12B & google/gemma-4-12B-it & Variable resolution; 16-pixel patches; max 280 soft tokens \\
GLM-4.6V-Flash & zai-org/GLM-4.6V-Flash & Dynamic resolution; 14-pixel patches; merge 2 \\
Qwen3.8-27B & Qwen/Qwen3.8-27B & Dynamic resolution; 16-pixel patches; merge 2 \\
Gemma4-31B & google/gemma-4-31B-it & Variable resolution; 16-pixel patches; max 280 soft tokens \\
DeepSeek-VL2-small & deepseek-ai/deepseek-vl2-small & 384-pixel dynamic tiles; 14-pixel patches; downsample 2 \\
DeepSeek-VL2 & deepseek-ai/deepseek-vl2 & 384-pixel dynamic tiles; 14-pixel patches; downsample 2 \\
\bottomrule
\end{tabularx}
\end{table}
Task 4 uses BF16 weights; all other tasks use NF4 double quantization with BF16 computation in quantized layers. VGL-stGen uses its merged model for task 4 and its deployed adapter for the other tasks. Gemma 4's multimodal projection and vision modules excluded by its loading configuration remain unquantized; k-bit preparation can cast unquantized parameters to FP32. DeepSeek uses eager language attention throughout and its native memory-efficient vision attention. Other models use SDPA for tasks 1–9 and eager attention for tasks 10–13. Model-dependent loading and tokenization adaptations preserve the requested input and score definition.

For categorical outputs, allowed-label tries are constructed with each tokenizer. Gene-list outputs are parsed using the original permissive gene parser. Expression completion fixes the 25 gene names and selects successive level tokens autoregressively; prefix-sharing tokens are handled as branches of the allowed-level trie. Model-specific end-of-turn markers delimit the answer, and complete raw generations are retained for audit. Structured result records are not counted as evidence that a model freely generated valid JSON.

For task 5, the same pooled marginal-probability correction is applied to each model. Raw and corrected macro-F1 are both reported. This is a transductive comparison on the development subset, and the correction was selected after labelled evaluation. For task 10, the requested gene-list length is the true answer length. Tasks 11 and 12 retain 75 observed levels; this benchmark is conditional completion rather than expression prediction from H\&E alone.

\section{Benchmark reproducibility and uncertainty}
Twelve of thirteen VGL-stGen aggregate scores match the historical task inventory at its reported precision. For task 4, the matched H100 rerun has observed balanced accuracy 0.7526 and a slide-bootstrap mean of 0.7524. The latter is the point estimate in the new model comparison. The earlier 0.7532 remains in the original task inventory and encoder comparison, representing that earlier evaluation. Gene-list point estimates preserve the original sample-weighted aggregation of shard-rounded scores; bootstrap statistics instead use unrounded per-row scores. Consequently, some paired differences differ in the last displayed digit from subtraction of the displayed table entries.

We resample the same units jointly for VGL-stGen and each comparator, using 1,000 draws and seed 0. Tasks 1–3 resample cells or spots, task 4 resamples slides, task 5 resamples patches, and tasks 6–13 resample input rows. Task 5 recomputes the pooled correction within each bootstrap draw. Gene sets and all queried levels within a row remain together. These intervals measure conditional sampling uncertainty for the fitted models. They do not include training variation, account for all section or patient clustering, or correct for the multiple comparisons. They are separate from the section-bootstrap intervals used for the earlier task-specific gene classifiers.

Table 12 reports VGL-stGen differences from the comparator with the highest point estimate for each task. This comparator is selected descriptively from the same results and can differ between rows. The ancillary tables provide point intervals and paired intervals for every comparator, including the format-constrained arm. Fig. 5 visualises the complete primary score matrix without averaging across metrics.

\begin{figure}[!htbp]
\centering
\includegraphics[width=122.000mm]{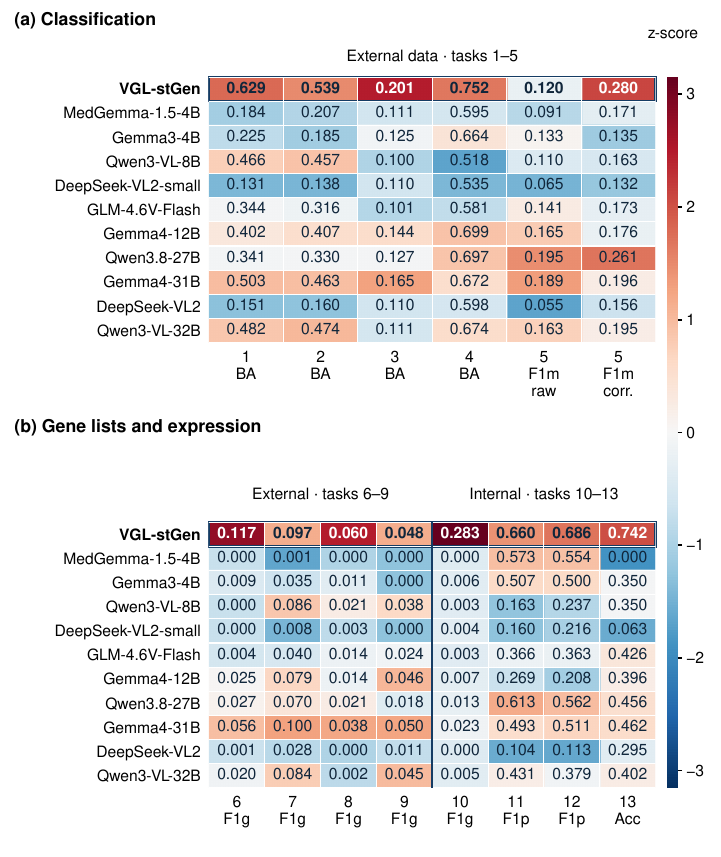}
\caption{Comparison with ten unadapted vision-language models: (a) classification; (b) gene and expression tasks. Printed values are original scores; colors are column-wise z-scores across 11 models, using population standard deviation. T5 raw and pooled-corrected columns are standardized separately. BA: balanced accuracy; F1m/F1g/F1p: macro/gene-set/pair-F1; Acc: accuracy. Task key: T1–3, cell typing from genes, image + genes, or image, respectively; T4, image \ensuremath{\rightarrow} binary tumor status; T5, image \ensuremath{\rightarrow} nine tissue classes; T6–7, gene names from image or cell type; T8–9, missing genes from partial lists plus image or cell type; T10, image \ensuremath{\rightarrow} Very High genes; T11–12, queried levels from observed levels plus image or cell type; T13, gene–level pairs \ensuremath{\rightarrow} cell type. For paired task IDs, inputs follow numerical order.}
\label{fig:vlm_overview}
\end{figure}
\begin{table}[!htbp]
\centering\small
\caption{Paired VGL-stGen minus highest-scoring comparator differences, with unadjusted 95\% bootstrap intervals (1,000 draws). T5 uses corrected F1m. Units: cells/spots (T1–3), slides (T4), patches (T5), complete rows (T6–13). Differences use unrounded underlying scores; T4 uses observed BA rather than the displayed bootstrap mean. Comparator selection is descriptive and not adjusted for selection or multiple comparisons. Intervals condition on fitted models, not independent-patient generalization. Task key: T1, genes \ensuremath{\rightarrow} cell type; T2, image + genes \ensuremath{\rightarrow} cell type; T3, image \ensuremath{\rightarrow} cell type; T4, image \ensuremath{\rightarrow} binary tumor status; T5, image \ensuremath{\rightarrow} nine-class tissue type; T6, image \ensuremath{\rightarrow} gene names; T7, cell type \ensuremath{\rightarrow} gene names; T8, image + partial gene list \ensuremath{\rightarrow} missing gene names; T9, cell type + partial gene list \ensuremath{\rightarrow} missing gene names; T10, image \ensuremath{\rightarrow} very-high-expression gene names; T11, image + observed levels \ensuremath{\rightarrow} queried levels; T12, cell type + observed levels \ensuremath{\rightarrow} queried levels; T13, gene–level pairs \ensuremath{\rightarrow} cell type.}
\label{tab:vlm_intervals}
\begin{tabularx}{\textwidth}{@{}>{\hsize=0.320\hsize\raggedright\arraybackslash}X>{\hsize=1.400\hsize\raggedright\arraybackslash}X>{\hsize=0.800\hsize\raggedright\arraybackslash}X>{\hsize=1.480\hsize\raggedright\arraybackslash}X@{}}
\toprule
Task & Comparator & Difference & 95\% interval \\
\midrule
1 & Gemma4-31B & +0.1258 & [+0.1204, +0.1309] \\
2 & Qwen3-VL-32B & +0.0654 & [+0.0598, +0.0709] \\
3 & Gemma4-31B & +0.0360 & [+0.0332, +0.0386] \\
4 & Gemma4-12B & +0.0538 & [+0.0491, +0.0587] \\
5 & Qwen3.8-27B & +0.0188 & [+0.0133, +0.0238] \\
6 & Gemma4-31B & +0.0617 & [+0.0568, +0.0665] \\
7 & Gemma4-31B & -0.0037 & [-0.0076, +0.0006] \\
8 & Gemma4-31B & +0.0222 & [+0.0154, +0.0292] \\
9 & Gemma4-31B & -0.0018 & [-0.0090, +0.0056] \\
10 & Gemma4-31B & +0.2599 & [+0.2456, +0.2741] \\
11 & Qwen3.8-27B & +0.0466 & [+0.0417, +0.0515] \\
12 & Qwen3.8-27B & +0.1236 & [+0.1164, +0.1308] \\
13 & Gemma4-31B & +0.2800 & [+0.2490, +0.3130] \\
\bottomrule
\end{tabularx}
\end{table}
\section{Format-constrained generation}
Table 13 repeats tasks 6–10 and 13 with output constraints, for all eleven models. Gene-list decoding enforces array syntax while permitting unrestricted gene-symbol strings made of letters, digits, periods and hyphens. It supplies neither a gene-panel whitelist nor the biological answer. Levels-to-cell-type decoding permits the printed, lowercase and uppercase spellings of each label, with case-insensitive scoring. Inputs, token limits, batches and metrics are unchanged. A maximum-token cutoff can still truncate an output, so this arm does not guarantee that every generation is complete JSON.

\begin{table}[!htbp]
\centering\scriptsize\setlength{\tabcolsep}{2pt}
\caption{Format-constrained generation on the same 1,000 rows per task, with unchanged NF4 precision, inputs and token limits. T6–10 report gene-set F1 under a gene-array grammar without a gene-panel whitelist; T13 reports accuracy under allowed-label spellings. Fixed token limits can still truncate outputs. Values preserve the original aggregation and rounding. Task key: T6, image \ensuremath{\rightarrow} gene names; T7, cell type \ensuremath{\rightarrow} gene names; T8, image + partial gene list \ensuremath{\rightarrow} missing gene names; T9, cell type + partial gene list \ensuremath{\rightarrow} missing gene names; T10, image \ensuremath{\rightarrow} very-high-expression gene names; T13, gene–level pairs \ensuremath{\rightarrow} cell type.}
\label{tab:vlm_forced}
\begin{tabularx}{\textwidth}{@{}>{\hsize=1.960\hsize\raggedright\arraybackslash}X>{\hsize=0.840\hsize\raggedright\arraybackslash}X>{\hsize=0.840\hsize\raggedright\arraybackslash}X>{\hsize=0.840\hsize\raggedright\arraybackslash}X>{\hsize=0.840\hsize\raggedright\arraybackslash}X>{\hsize=0.840\hsize\raggedright\arraybackslash}X>{\hsize=0.840\hsize\raggedright\arraybackslash}X@{}}
\toprule
Model & T6 & T7 & T8 & T9 & T10 & T13 \\
\midrule
VGL-stGen & 0.1173 & 0.0965 & 0.0604 & 0.0484 & 0.2828 & 0.7420 \\
MedGemma-1.5-4B & 0.0005 & 0.0140 & 0.0065 & 0.0141 & 0.0011 & 0.3620 \\
Gemma3-4B & 0.0186 & 0.0625 & 0.0088 & 0.0180 & 0.0057 & 0.3530 \\
Qwen3-VL-8B & 0.0024 & 0.0850 & 0.0210 & 0.0380 & 0.0032 & 0.3500 \\
DeepSeek-VL2-small & 0.0000 & 0.0103 & 0.0020 & 0.0028 & 0.0043 & 0.0630 \\
GLM-4.6V-Flash & 0.0082 & 0.0501 & 0.0135 & 0.0262 & 0.0040 & 0.4220 \\
Gemma4-12B & 0.0497 & 0.0810 & 0.0300 & 0.0453 & 0.0244 & 0.3980 \\
Qwen3.8-27B & 0.0237 & 0.0699 & 0.0198 & 0.0456 & 0.0118 & 0.4560 \\
Gemma4-31B & 0.0611 & 0.0995 & 0.0392 & 0.0508 & 0.0230 & 0.4720 \\
DeepSeek-VL2 & 0.0005 & 0.0281 & 0.0000 & 0.0131 & 0.0000 & 0.2950 \\
Qwen3-VL-32B & 0.0197 & 0.0839 & 0.0022 & 0.0452 & 0.0047 & 0.4020 \\
\bottomrule
\end{tabularx}
\end{table}
VGL-stGen retains its reported scores. Several comparators improve, including MedGemma's levels-to-cell-type accuracy from 0.0000 to 0.3620 and Gemma4-31B's from 0.4620 to 0.4720. VGL-stGen remains at 0.7420; its paired difference from Gemma4-31B is 0.2700, with 95\% interval [0.2380, 0.3020]. The two cell-type-conditioned gene-list tasks remain close to Gemma4-31B. Thus, enforcing output format does not remove the principal image-gene and cell-typing gaps, but the comparison remains conditional on the fixed generation budgets.

Inspection of low-scoring MedGemma outputs shows that intermediate reasoning often consumes the available generation budget. Parsed free-list outputs can also name genes outside the evaluation panel or use aliases not mapped by the scorer. Out-of-panel genes are not necessarily fictitious. These observations explain why a near-zero score under this protocol should not be read as absence of biological knowledge. Under format constraints, MedGemma predicts T cells in 999 of 1,000 levels-to-cell-type rows, yielding the majority-label score. A supplementary diagnostic on the first 100 rows uses full-label likelihood scoring, rather than the primary decoder: removing the molecular input preserves a constant prediction, whereas reversing candidate order changes that prediction. This diagnostic identifies sensitivity to the readout and candidate ordering without establishing a biological mechanism for the bias.

\section{Supporting analyses of learning and inputs}
\subsection{Expression-adaptation design}
Fig. 6 contrasts full-answer and level-only supervision. The optimisation settings, development selection and image-permutation protocol are specified in Detailed methods under Controlled expression adaptation and image permutation. The comparison holds inputs and autoregressive evaluation fixed.

\begin{figure}[!htbp]
\centering
\includegraphics[width=122.000mm]{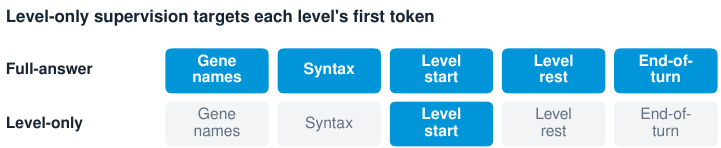}
\caption{Supervised answer positions in the separate expression-adaptation study. Blue groups receive full-vocabulary cross-entropy. Full-answer supervision includes all answer tokens and end-of-turn; level-only supervision selects each queried level's first token. Level rest denotes any remaining level tokens. Initial weights, inputs, update budget and autoregressive evaluation are matched. These checkpoints are separate from the deployed 13-task model.}
\label{fig:expression_supervision}
\end{figure}
\subsection{Full-input comparators and uncertainty}
Comparator architectures and fitting procedures are specified in Detailed methods under Context-aware expression baselines, Gene classifiers and adaptation-capacity control, and Image encoders and within-cohort reference fits. Bootstrap procedures are specified under Statistical analysis.

\subsection{Gene-mediated image supervision}
Image-to-cell-type balanced accuracy increases from 0.161 to 0.226 after self-distillation and is 0.2012 after cancer training and averaging. On the probes' common 54,058-item subset, deployed VGL-stGen scores 0.1998 and transferred probes reach at most 0.195; target-cohort probe fits span 0.27–0.35. Joint image-and-gene prompting scores 0.5394, below genes alone at 0.6297 in the same evaluation. The joint task received no direct supervision.

\subsection{Conditional expression completion and adaptation}
Fixing query names separates output formatting from level prediction. The parser identifies all masked names across 176,008 held-out rows. On 1,000 rows per task, free-generation pair-F1 is 0.0361 (image) and 0.0123 (cell type); fixed-name autoregressive decoding gives 0.6596 and 0.6858 with unchanged weights. The image task exceeds its per-gene frequency table, 0.6198, by +0.0398 [0.0350, 0.0445]. The cell-type task trails its conditional table, 0.6963, by \ensuremath{-}0.0106 [\ensuremath{-}0.0160, \ensuremath{-}0.0052]. These tables omit visible levels. At 75 observed levels, full-input MLP references score higher on both tasks (Table 14).

\begin{table}[!htbp]
\centering\small
\caption{Expression pair-F1 on the same 1,000 rows per task. Additional training and MLPs: mean \ensuremath{\pm} SD of 3 individual seeds, not ensembles. Each VGL-stGen adaptation uses 4,096 additional rows and 256 updates. MLPs use original task training splits; the image MLP uses UNI2-h, so encoders and training histories differ.}
\label{tab:adaptation}
\begin{tabularx}{\textwidth}{@{}>{\hsize=1.260\hsize\raggedright\arraybackslash}X>{\hsize=0.870\hsize\raggedright\arraybackslash}X>{\hsize=0.870\hsize\raggedright\arraybackslash}X@{}}
\toprule
Model / supervision & Image & Cell type \\
\midrule
Deployed checkpoint & 0.6596 & 0.6858 \\
Full-answer adaptation & 0.6720 \ensuremath{\pm} 0.0009 & 0.7221 \ensuremath{\pm} 0.0007 \\
Level-only adaptation & 0.6805 \ensuremath{\pm} 0.0011 & 0.7201 \ensuremath{\pm} 0.0004 \\
Full-input MLP & 0.6811 \ensuremath{\pm} 0.0038 & 0.7300 \ensuremath{\pm} 0.0008 \\
\bottomrule
\end{tabularx}
\end{table}
With decoding held fixed, both additional-training conditions improve both tasks in all three seeds. Level-only exceeds full-answer supervision on the image task by +0.00848, sample CI [0.00561, 0.01015]; the direction is positive in every seed on both platforms (Fig. 7). Its cell-type difference is \ensuremath{-}0.00200, row CI [\ensuremath{-}0.00507, 0.00113].

\begin{figure}[!htbp]
\centering
\includegraphics[width=122.000mm]{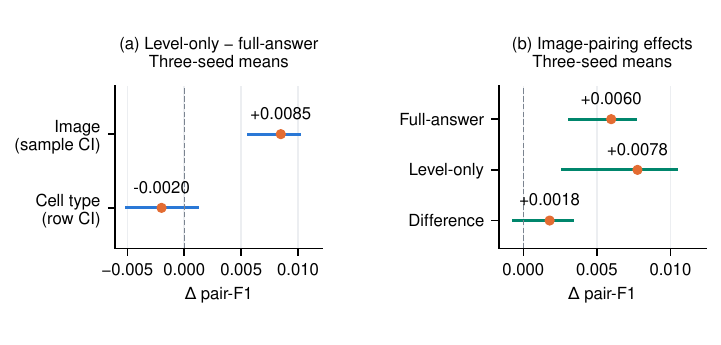}
\caption{Controlled expression prediction after additional adaptation. (a) Level-only minus full-answer pair-F1 with intact inputs. (b) Intact-minus-permuted image effects; Difference denotes the level-only effect minus the full-answer effect. Points average three matched seeds. Conditional 95\% bootstrap intervals resample image samples or cell-type rows and condition on fitted models. Panel b resamples samples. Image intervals preserve platform proportions. The 25 queries and within-sample/platform permutation are fixed. Sample identifiers are not verified patients.}
\label{fig:expression}
\end{figure}
Absent targets comprise 61.36\% (image) and 58.92\% (cell type). Mean non-Absent accuracy rises from full-answer to level-only supervision: 0.2664 to 0.2930 (image) and 0.4433 to 0.4535 (cell type). MLPs remain higher at 0.3160 and 0.4786. Thus, the cell-type trade-off depends on the metric.

Across three training seeds, intact-minus-permuted pair-F1 is +0.00597 [0.00316, 0.00764] for full-answer and +0.00776 [0.00263, 0.01042] for level-only supervision (sample CIs). On the pooled image benchmark, per-seed differences are positive in both conditions. The between-objective difference is +0.00179 [\ensuremath{-}0.00067, 0.00336], leaving an increase in the image-pairing effect unresolved.

For the fixed level-only seed-42 checkpoint, reducing observed levels from 75 to 25 lowers image pair-F1 from 0.68032 to 0.65992. The change in image-pairing effect remains unresolved. The extended methods report all budgets and secondary metrics under Completion with fewer observed expression levels.

\subsection{Matched gene-readout control}
Table 15 gives the matched gene-task comparison and trainable parameter counts.

\begin{table}[!htbp]
\centering\small
\caption{Matched adaptation-capacity control on the gene task. These short runs are separate from deployment. Data, rank, learning rate, schedule and update budget are matched; output heads and losses differ.}
\label{tab:controls}
\begin{tabularx}{\textwidth}{@{}>{\hsize=1.600\hsize\raggedright\arraybackslash}X>{\hsize=0.800\hsize\raggedright\arraybackslash}X>{\hsize=0.800\hsize\raggedright\arraybackslash}X>{\hsize=0.800\hsize\raggedright\arraybackslash}X@{}}
\toprule
Readout & Balanced accuracy & Macro-F1 & Trainable parameters \\
\midrule
Discriminative short run & 0.5554 & 0.5209 & 32,819,212 \\
Generative short run & 0.5737 & 0.5683 & 32,788,480 \\
\bottomrule
\end{tabularx}
\end{table}
On the gene task, short runs with matched data, adaptation rank and update budget give macro-F1 0.5683 for the generative readout and 0.5209 for the discriminative readout, a difference of 0.0474. Both the output head and loss differ; these single runs are separate from deployment.

\subsection{Transfer and task retention}
External gene-list tasks (Table 16) exceed transferred global-frequency F1 (0.0367 for generation; 0.0098 for completion) but trail target-cohort cell-type-conditional F1 (0.4009 and 0.2300, respectively). Training and external gene-frequency distributions differ; their top-twenty lists share five members.

From the pre-cancer checkpoint to deployment, genes-to-cell-type changes by \ensuremath{-}0.0079 macro-F1 with section CI [\ensuremath{-}0.0207, +0.0234], and image-to-cell-type changes by \ensuremath{-}0.0248 balanced accuracy. These changes cover cancer training, replay and averaging.

\subsection{Reference fits and label definitions}
Distribution-shift studies motivate comparing transferred performance with within-cohort fits \citep{r02,r03,r04,r05}. In this study, the range of transferred scores is 2.09 times the range of target-cohort scores across nine gene comparators and 2.17 times that range across fourteen cancer encoders. Target-cohort fits use additional labels, with cell-level folds for genes and patient-level folds for cancer. These ratios describe the selected comparator panels and evaluation designs.

On the same 452 slides, refitting each cancer probe under the wide (TUMOR, NECROSIS and MIXED counted as positive) and TUMOR-only mappings yields a median paired difference of 0.1191 balanced accuracy, compared with 0.0436 for the colour reference. Each encoder's patient-paired interval excludes zero. Target definition, prevalence and probe fitting change together. Because the mapping was chosen during development, this is an exploratory measure of benchmark sensitivity.

Patient-level resampling of the binary VGL-stGen result gives an interval of [0.7441, 0.7622], compared with the slide interval [0.7452, 0.7610]. This sensitivity analysis accounts for repeated slides from a patient. Annotation error and adaptive development remain outside both intervals. Additional comparisons and implementation details appear in the supplementary methods and figures.

\section{Region-query application example}
VGL-stGen outputs can be organised for region-level tissue inspection and spatial molecular analysis. Fig. 8 reproduces original application examples of a region-query workflow: a user draws a box on a slide and asks, "score this region for cancer and show me the map." The surrounding application selects tissue, runs image-conditioned predictions and displays the scores alongside H\&E.

\begin{figure}[!htbp]
\centering
\includegraphics[width=160.000mm]{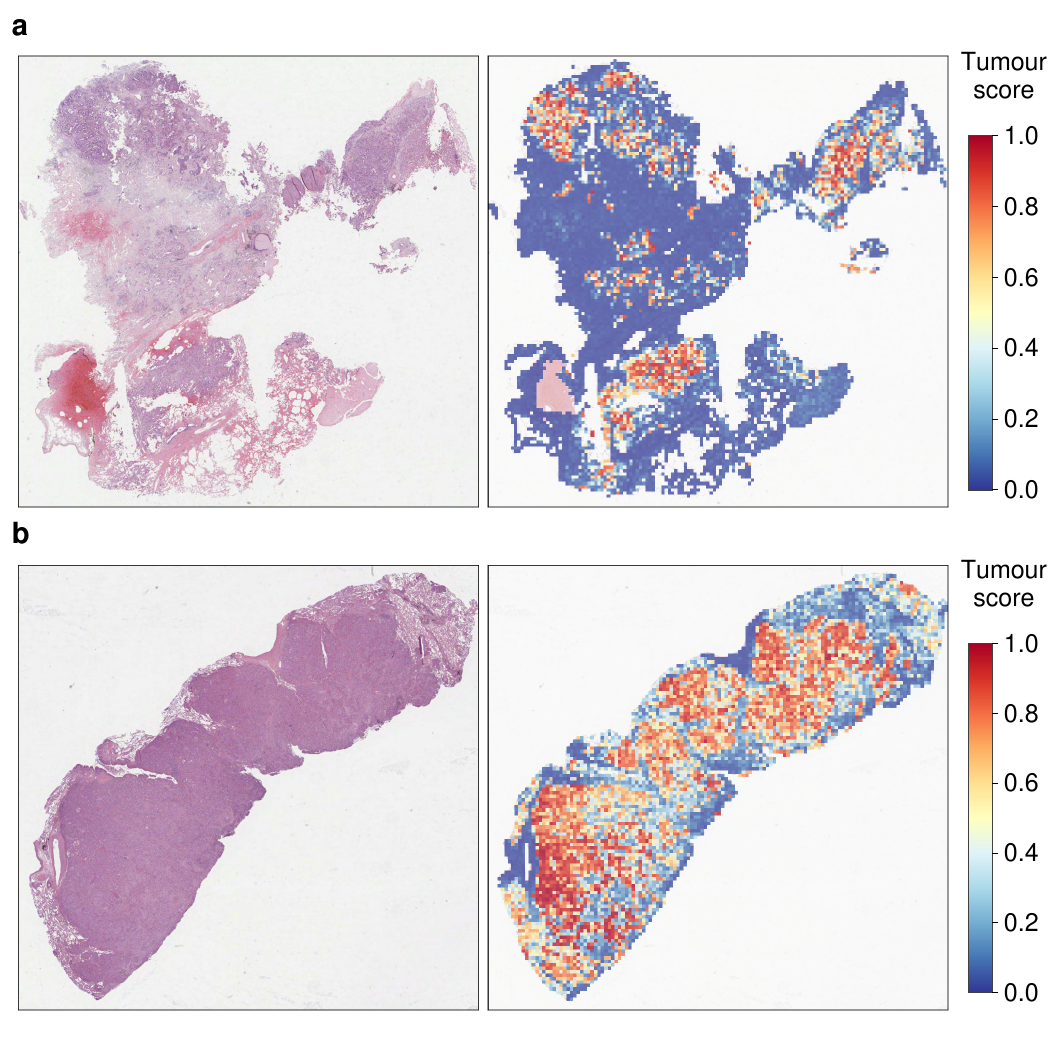}
\caption{Region-query cancer mapping. H\&E tissue (left) and the corresponding model cancer-score map (right) are shown for (a) TCGA-73-7499-01Z-00-DX1 and (b) TCGA-55-6979-01Z-00-DX1. These application examples illustrate selecting tissue and requesting a cancer-score map. Blue and red indicate lower and higher model scores, respectively. Scores are model outputs, not calibrated clinical probabilities.}
\label{fig:ppt_roi}
\end{figure}
\section{Additional comparison panels}
Fig. 9 maps the served tasks to their output types and recorded evaluation conditions. Fig. 10 locates the reported checkpoints in the training history.

\begin{figure}[!htbp]
\centering
\includegraphics[width=122.000mm]{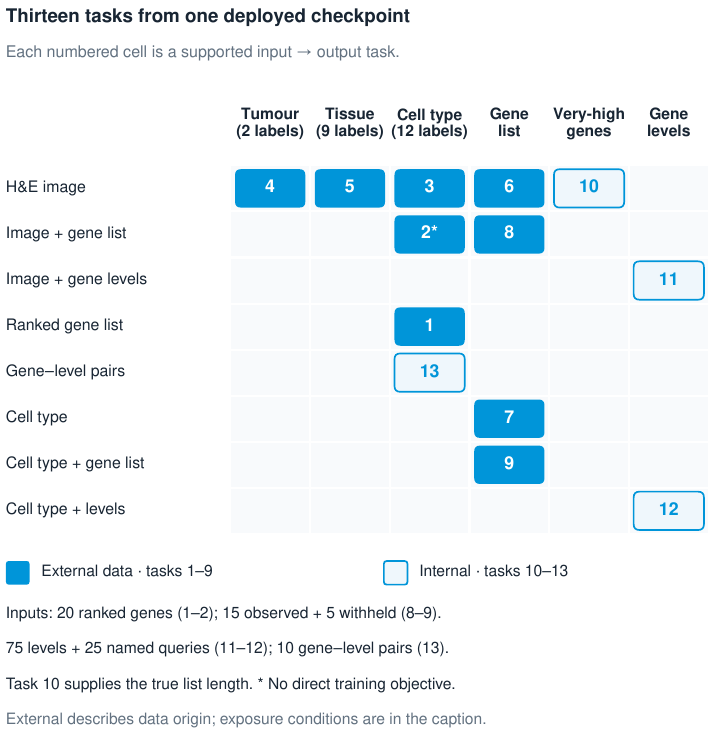}
\caption{Input–output map of the 13 deployed tasks. Numbers match the task inventory; the asterisk marks task 2 without a direct training objective. Filled cells use external datasets; outlined cells use internal evaluation data. External refers to data origin: 3 spatial sources supplied cancer-training H\&E, task 4 includes TCGA development slides, and task 5 uses the development pool. Classification references are computationally derived. Gene-list completion supplies 15 names and predicts five; expression completion supplies 75 levels and queries 25 named genes. Task 10 supplies true answer length; task 13 supplies ten gene–level pairs. All tasks share the same checkpoint.}
\label{fig:detail1}
\end{figure}
\begin{figure}[!htbp]
\centering
\includegraphics[width=160.000mm]{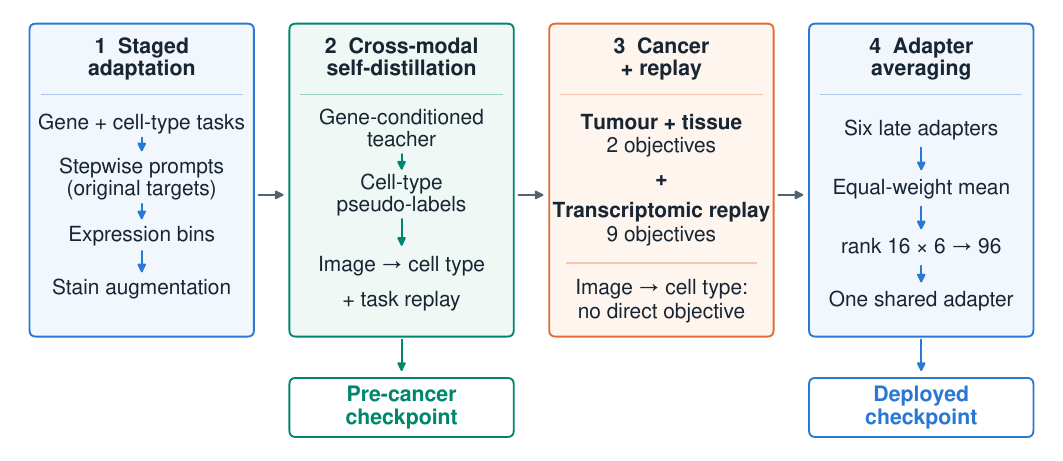}
\caption{Training stages and evaluated checkpoints. Gene and cell-type adaptation precedes gene-mediated image supervision, joint cancer training with replay, and averaging. Stepwise prompts retain the original targets. The vision tower remains frozen. The pre-cancer checkpoint follows image supervision; deployment follows averaging. The averaging window was selected using the 39-slide TCGA development subset.}
\label{fig:detail2}
\end{figure}
Fig. 11 reports both gene-classification metrics and section-paired differences. Fig. 12 gives the matched adaptation-capacity comparison. Fig. 13 contrasts transferred and target-cohort probe scores on separate axes. Fig. 14 shows the gene-budget dependence of a logistic reference.

\begin{figure}[!htbp]
\centering
\includegraphics[width=160.000mm]{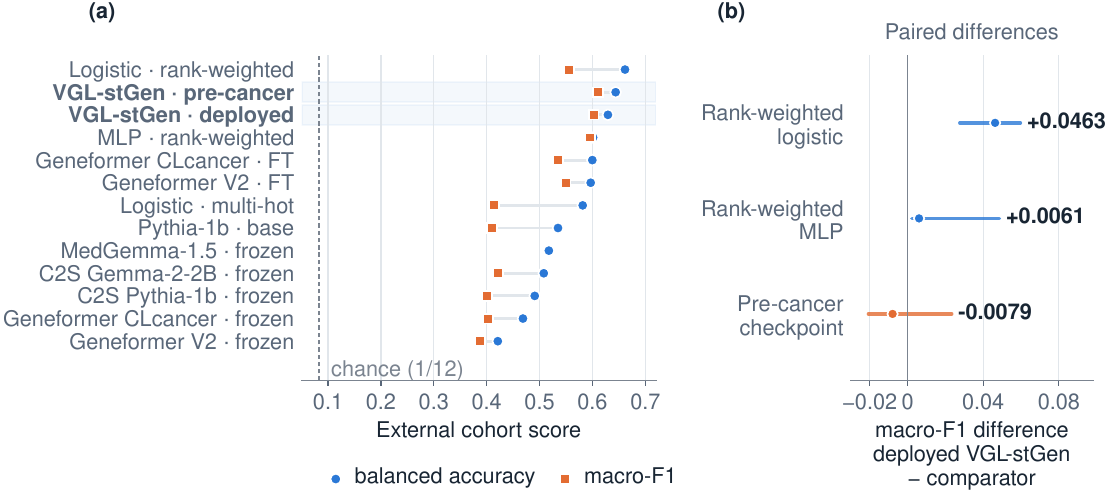}
\caption{(a) Gene classifiers and VGL-stGen checkpoints on 224,210 items. The pre-cancer point uses its original evaluation (macro-F1 0.6104). FT denotes fine-tuning. Macro-F1 was not scored for the frozen-representation MedGemma probe in this panel; that probe is separate from the direct multimodal-model benchmark. (b) Paired macro-F1 differences with section-bootstrap intervals. The pre-cancer score in the paired comparison is 0.6102, versus 0.6023 after joint training and averaging. The MLP comparison is provisional because of discrepancies between refits.}
\label{fig:detail3}
\end{figure}
\begin{figure}[!htbp]
\centering
\includegraphics[width=95.791mm]{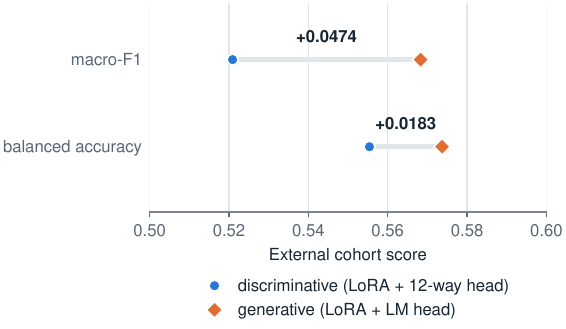}
\caption{Generative and discriminative readouts for gene-based cell typing under matched adaptation settings: rank, learning rate, schedule, batch, 6,000 updates, seed, prompts and data (40,000 evaluation items). Output heads and losses differ. Parameter counts are reported in the detailed Methods. Points are single-run estimates, separate from the deployed checkpoint.}
\label{fig:detail4}
\end{figure}
\begin{figure}[!htbp]
\centering
\includegraphics[width=160.000mm]{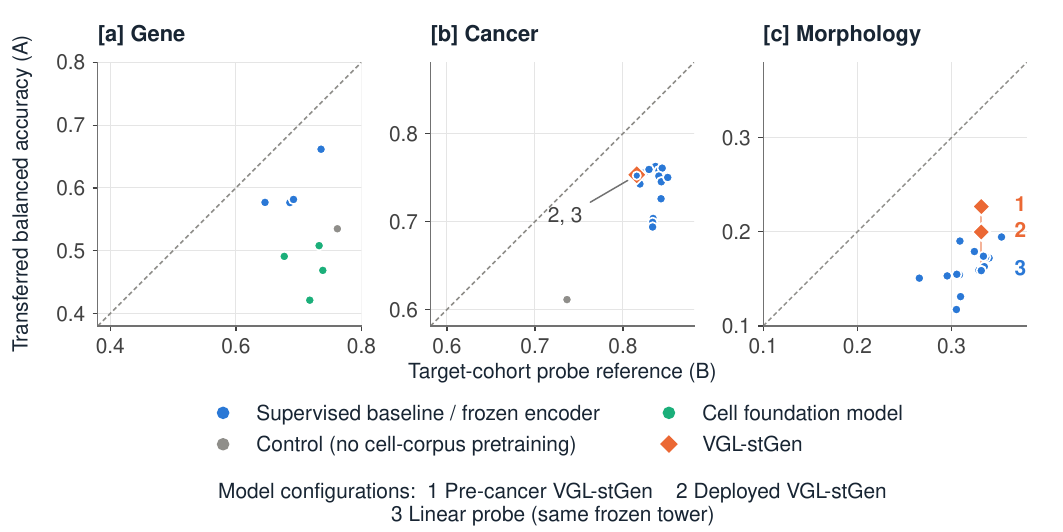}
\caption{Transferred versus target-cohort probe scores for (a) gene, (b) cancer and (c) morphology classification. Both axes report balanced accuracy; the diagonal denotes equal scores. Target-cohort fits use cell-level folds for genes and morphology and patient-level folds for cancer. The numbered model configurations are 1, pre-cancer VGL-stGen; 2, deployed VGL-stGen; 3, a task-specific linear probe on the same frozen vision tower. Panel (b) shows 2 and 3; panel (c) shows all three. All morphology comparisons use the same 54,058 evaluation items. VGL-stGen imaging coordinates use its own-tower probe reference, not a fitted VGL-stGen upper bound. Classification references are silver; spatial H\&E exposure is partial and TCGA includes development slides.}
\label{fig:detail5}
\end{figure}
\begin{figure}[!htbp]
\centering
\includegraphics[width=85.542mm]{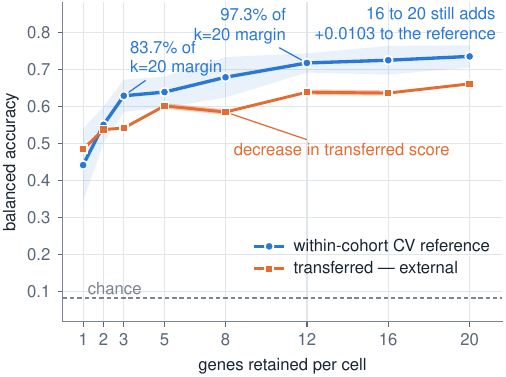}
\caption{Gene-budget analysis for logistic regression: target-cohort cross-validation and transferred balanced accuracy as the input list is shortened. Shading denotes fold standard deviation. Percentages refer to the target-cohort score's margin above chance at 20 genes.}
\label{fig:detail6}
\end{figure}
Fig. 15 gives the cancer comparison on 452 slides. Fig. 16 compares it with the earlier 39-slide subset, and Fig. 17 shows sensitivity to cancer-label definitions. Fig. 18 compares training and external gene frequencies.

\begin{figure}[!htbp]
\centering
\includegraphics[width=137.000mm]{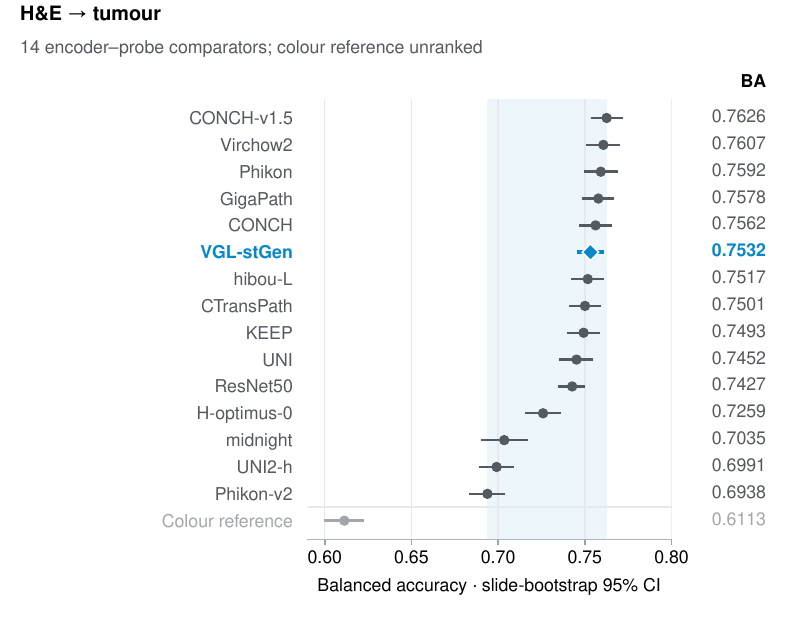}
\caption{TCGA-LUAD tumor prediction: VGL-stGen and 14 encoder–probe systems, including the ImageNet ResNet50 reference. The color-histogram reference is unranked. Scores measure agreement with ARA-CNN-derived tumor labels; intervals resample slides. The shaded band spans encoder point estimates, not confidence bounds. The 452-slide set includes the 39-slide development subset; rank is descriptive.}
\label{fig:detail7}
\end{figure}
\begin{figure}[!htbp]
\centering
\includegraphics[width=88.208mm]{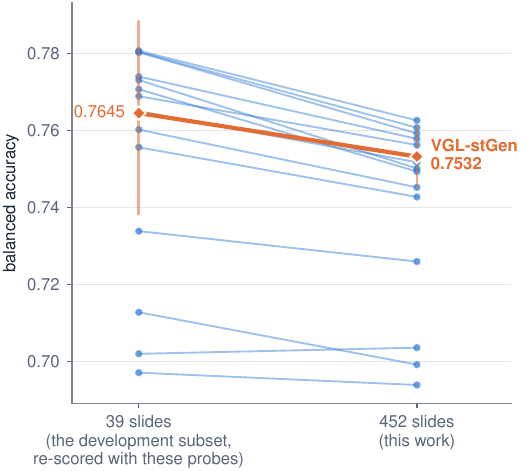}
\caption{Cancer scores on the earlier 39-slide subset and the expanded 452-slide evaluation. 13 of 14 encoders and VGL-stGen have lower point estimates on the expanded set. Slide-bootstrap intervals narrow with the larger sample; the two sets overlap and are not independent replications.}
\label{fig:detail8}
\end{figure}
\begin{figure}[!htbp]
\centering
\includegraphics[width=103.648mm]{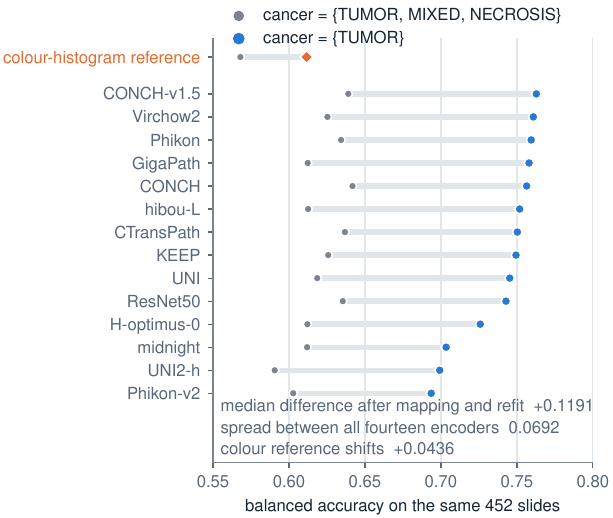}
\caption{Cancer-label sensitivity on the same 452 slides. Each encoder probe is refitted under the wide mapping (TUMOR, NECROSIS and MIXED positive) and the TUMOR-only mapping. Grey and blue indicate the respective mappings; orange marks the colour reference under TUMOR-only. Targets, prevalence and probe fits change together.}
\label{fig:detail9}
\end{figure}
\begin{figure}[!htbp]
\centering
\includegraphics[width=103.000mm]{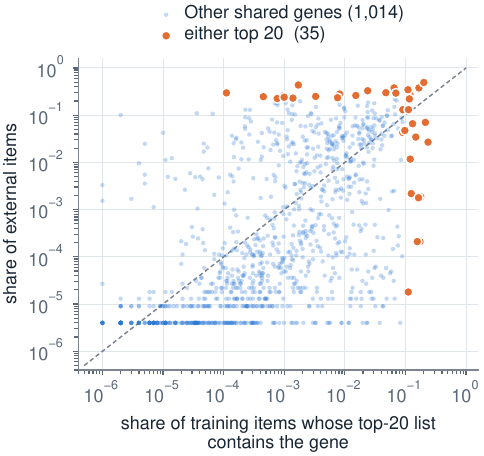}
\caption{Training and external gene frequencies across 1,049 shared symbols. Blue marks 1,014 genes in neither top-20 list; orange marks their 35-gene union. The dashed line denotes equal frequency. Spearman rho is 0.63. The five genes in both top-20 lists are ACTA2, EPCAM, ICAM1, MET and S100A9. Frequencies count cells or spots rather than task rows. Zero frequencies are placed at half the smallest positive observed frequency on the log axes.}
\label{fig:detail10}
\end{figure}
\section{Deployed task inventory}
Table 16 lists the served tasks and their evaluation conditions.

\begin{table}[!htbp]
\centering\small
\caption{Deployed-checkpoint task inventory. BA: balanced accuracy; F1m: macro-F1; F1g: gene-set F1; F1p: pair-F1; Acc: accuracy; AR: autoregressive. H\&E overlap denotes three of four spatial sources used in cancer training; SCC is absent from that manifest. Task 4 includes development slides; task 5 pools NF4-adapter scores from 30,719 patches / 39 slides for transductive correction, selected after labelled evaluation. Task 10 supplies the true answer length. Tasks 10/13 use baselines estimated from evaluation data; task 13 uses ten sampled gene–level pairs. Gene-list completion supplies 15 names and predicts five. Additional-training results are separate.}
\label{tab:inventory}
\begin{tabularx}{\textwidth}{@{}>{\hsize=0.200\hsize\raggedright\arraybackslash}X>{\hsize=1.800\hsize\raggedright\arraybackslash}X>{\hsize=1.150\hsize\raggedright\arraybackslash}X>{\hsize=0.600\hsize\raggedright\arraybackslash}X>{\hsize=1.250\hsize\raggedright\arraybackslash}X@{}}
\toprule
ID & Task & Score & Metric & Condition \\
\midrule
1 & Genes \ensuremath{\rightarrow} cell type & 0.6293 / 0.6023 & BA / F1m & External \\
2 & Image + genes \ensuremath{\rightarrow} cell type & 0.5394 & BA & H\&E overlap \\
3 & Image \ensuremath{\rightarrow} cell type & 0.2012 & BA & H\&E overlap \\
4 & Image \ensuremath{\rightarrow} tumour & 0.7532 & BA & 452 slides \\
5 & Image \ensuremath{\rightarrow} tissue & 0.1200 \ensuremath{\rightarrow} 0.2801 & F1m & 39-slide pool \\
6 & Image \ensuremath{\rightarrow} genes & 0.1173 & F1g & H\&E overlap \\
7 & Cell type \ensuremath{\rightarrow} genes & 0.0965 & F1g & External \\
8 & Image + partial \ensuremath{\rightarrow} genes & 0.0604 & F1g & H\&E overlap \\
9 & Cell type + partial \ensuremath{\rightarrow} genes & 0.0484 & F1g & External \\
10 & Image \ensuremath{\rightarrow} very-high genes & 0.2828 & F1g & Internal; true length \\
11 & Image + levels \ensuremath{\rightarrow} levels & 0.6596 & F1p & Internal AR \\
12 & Cell type + levels \ensuremath{\rightarrow} levels & 0.6858 & F1p & Internal AR \\
13 & Levels \ensuremath{\rightarrow} cell type & 0.7420 & Acc & Internal \\
\bottomrule
\end{tabularx}
\end{table}
\label{mainend}

\end{document}